\documentclass{article}

\PassOptionsToPackage{numbers,compress}{natbib}

\usepackage[main, final]{neurips_2026}

\usepackage[utf8]{inputenc}
\usepackage[T1]{fontenc}
\usepackage[table]{xcolor}
\usepackage{algorithm}
\usepackage{algpseudocode}
\algrenewcommand\algorithmiccomment[1]{\hfill{\color{green!45!black}$\triangleright$ #1}}
\usepackage{hyperref}
\usepackage{url}
\usepackage{booktabs}
\usepackage{amsfonts}
\usepackage{amsmath}
\usepackage{amssymb}
\usepackage{bm}
\usepackage{graphicx}
\usepackage{wrapfig}
\usepackage{multirow}
\usepackage{nicefrac}
\usepackage{microtype}
\newsavebox{\baselineTableBox}
\graphicspath{{figure/}}

\title{SRL-MPC: Shape-Aware Reinforcement Learned Model Predictive Control}

\author{
  Ruihua Han$^{1}$ \quad Rui Gao$^{2}$ \quad Zhe Liu$^{1}$ \quad
  Xinyi Wang$^{3}$ \quad Chang Chen$^{1}$ \\
  \textbf{Shuai Wang}$^{4}$ \quad \textbf{Qi Hao}$^{2}$ \quad
  \textbf{Jia Pan}$^{1}$ \quad \textbf{Hengshuang Zhao}$^{\dagger 1}$ \\ \\
  $^{1}$The University of Hong Kong \\
  $^{2}$Research Institute of Trustworthy Autonomous Systems, \\
  Southern University of Science and Technology \\
  $^{3}$University of Michigan \quad
  $^{4}$Shenzhen Institutes of Advanced Technology \\[0.5em]
  \textbf{Project:} \url{https://hku-sail.github.io/srl_mpc}
}

\hypersetup{
  pdftitle={SRL-MPC: Shape-Aware Reinforcement Learned Model Predictive Control},
  pdfauthor={Ruihua Han, Rui Gao, Zhe Liu, Xinyi Wang, Chang Chen, Shuai Wang, Qi Hao, Jia Pan, and Hengshuang Zhao}
}

\begin{document}

\maketitle
\begingroup
\renewcommand{\thefootnote}{\fnsymbol{footnote}}
\footnotetext[2]{Corresponding author.}
\endgroup

\begin{abstract}
Safe and efficient shape-aware navigation in heterogeneous crowds and robot fleets remains challenging. Traditional approaches often assume homogeneous robots, sparse workspaces, simplified geometry, offline computation, or handcrafted parameters to make the problem tractable, which limits their deployment in dense crowd scenarios. Toward this end, we propose Shape-Aware Reinforcement Learned Model Predictive Control (\textbf{SRL-MPC}), a method for efficient and adaptive navigation in crowds with heterogeneous shapes using explicit footprint models. To encourage shape-aware collision avoidance, we formulate high-order control barrier function (HOCBF) constraints from geometric separation features (GSFs) based on support function transformation. A reinforcement learning (RL) framework then learns a neural policy that reads GSFs and outputs real-time MPC parameter updates, enabling the MPC solver to adapt to neighboring crowd geometries. The key advantage of SRL-MPC is that it retains the explicit geometric and kinematic structure and generalizability of MPC while integrating the adaptability and intelligence of RL. Experiments in randomized crowd scenarios, out-of-distribution settings, and real-robot trials evaluate the performance of SRL-MPC. The results show dramatically improved success rates in dense scenarios through our combination of learned parameter adaptation and shape-aware HOCBF residual penalties. 
\end{abstract}

\section{Introduction}
\label{sec:intro}
Safe navigation in crowds with heterogeneous shapes is a central capability for robots operating in shared spaces with other robots and humans~\citep{francis2025principles}. Here, shape-aware navigation refers to explicitly incorporating occupied body geometry into motion planning and control. The problem becomes difficult when a large number of robots must satisfy safety, efficiency, and kinematic constraints simultaneously. Existing approaches, including velocity obstacle (VO)-based methods~\citep{van2011reciprocal, guo2021vr, martinez2025avocado}, optimization-based methods~\citep{zhang2020optimization, luo2020multi, li2024edge, li2025efficient}, and reinforcement learning (RL)-based methods~\citep{chen2019crowd, han2020cooperative, wang2024navformer, xu2025navrl}, have shown promising results, but they often rely on assumptions such as homogeneous robots, sparse workspaces, or simplified circular shape representations~\citep{van2011reciprocal, chen2019crowd, han2022reinforcement, singamaneni2024survey}, as well as offline computation or repeatedly tuned parameters for specific scenarios~\citep{zhang2020optimization, li2025efficient, kim2025learning}. These assumptions make the problem tractable, but they also limit deployment in dense and dynamic crowds, where robots may get stuck or collide if the geometric information is not considered explicitly.

The core difficulty is to avoid collisions with dense, dynamic, and arbitrarily shaped objects in real time. RL-based methods can learn strong neural policies for specific scenarios, but they are prone to overfitting and may not generalize to arbitrary shapes or scale to large crowds~\citep{singamaneni2024survey}. Optimization-based approaches have the ability to handle static shaped obstacles by explicitly formulating constraints using compact sets~\citep{zhang2020optimization, han2023rda} or building a prior grid map~\citep{rosmann2017kinodynamic}. Recent works highlight the promise of combining learning with optimization to integrate these advantages, such as embedding differentiable model predictive control (MPC) into actor-critic learning~\citep{romero2025actor, wang2026diffcvar}, and using learned dynamics inside sampling based MPC for agile adaptive control~\citep{xiao2025anycar}. However, these methods mainly focus on single robot agile control, and their safety properties typically depend on the accuracy of learned costs or dynamics rather than explicit shape-aware safety terms.

To this end, this paper proposes \texttt{SRL-MPC}, a distributed shape-aware reinforcement learned Model Predictive Control (MPC) framework for robot navigation in crowds of robots and obstacles with explicitly modeled footprints. The key idea is to formulate a safe set based on geometric separation features (GSFs) as high-order control barrier function (HOCBF) constraints in the primal optimization problem for collision avoidance, while using RL to adjust MPC parameters based on the neighboring agent GSFs for task progress. After problem decomposition, the local MPC subproblem handles the HOCBF condition through a soft shape-aware residual penalty. In this way, RL does not replace the model-based planner; instead, it adapts the parameters of an explicit shape-aware HOCBF-MPC problem to balance collision avoidance and task progress. This approach has several advantages. First, unlike RL-based approaches that are sensitive to carefully designed reward functions and may suffer from limited generalization~\citep{han2022reinforcement, xu2025navrl}, the proposed method learns parameter adaptation for an explicit shape-aware HOCBF-MPC problem rather than an unconstrained end-to-end policy. Second, the HOCBF-MPC parameters are adapted from neighboring GSFs, avoiding repeated manual weight tuning for different scenarios as required by many optimization-based approaches~\citep{li2025efficient}. Third, the proposed method directly handles explicit convex geometric shape representations. For nonconvex footprints, the current implementation computes separation directions from the original geometry. Finally, following the problem decomposition technique in~\citep{han2025neupan}, the deployed local HOCBF-MPC subproblem is simple enough to solve in real time for each robot, resulting in a practical solution for dense dynamic crowd navigation.

To highlight the effectiveness of \texttt{SRL-MPC}, we evaluate it in highly randomized and dense scenarios consisting of multiple differential-drive robots, where the positions, goals, and shapes are all randomly generated. Additional evaluations cover out-of-distribution scenarios, variation across training seeds, and real-robot trials. 

\section{Related Work}
\label{sec:related_work}
\textbf{Traditional Approaches.}
Traditional collision avoidance approaches often rely on geometric or optimization-based formulations. Recent VO-based methods such as Adaptive Optimal Collision Avoidance Driven by Opinion (\texttt{AVOCADO}) estimate an agent's cooperation level online through nonlinear opinion dynamics, improving collision avoidance in mixed crowds without communication~\citep{martinez2025avocado}. MPC is a popular optimization framework that optimizes controls over a receding horizon with various constraints~\citep{zeng2021safety, chen2023multi}. To improve shape-aware collision avoidance, prior methods introduce disk primitives~\citep{ziegler2010fast}, polytopic velocity obstacles~\citep{huang2023velocity}, sequential convex optimization~\citep{schulman2014motion}, dual optimization-based collision avoidance (OBCA) constraints for convex sets~\citep{zhang2020optimization}, and accelerated optimization by problem decomposition and edge computation~\citep{han2023rda,li2024edge}. These methods improve geometric fidelity, but their exact constraints can grow with the number of object surfaces. \texttt{SRL-MPC} follows this optimization line by using GSFs derived from a support function representation, a compact fixed-dimensional representation to encode convex objects without scaling with the number of surfaces.

\textbf{Reinforcement Learning Approaches.}
RL approaches learn navigation policies from interaction data and have become an important category in socially aware and crowd-aware robot navigation~\citep{long2018towards, fan2020distributed, singamaneni2024survey}. Early socially aware deep reinforcement learning (DRL) methods learn collision-avoidance policies from local observations and social rewards~\citep{chen2017socially, everett2021collision, wang2022adaptive}. Typical methods such as socially aware reinforcement learning (\texttt{SARL}) use attention mechanisms to encode human-robot interactions in a crowd-level representation~\citep{chen2019crowd}. More recent methods use richer sequence, occupancy-map, or transformer-based representations to reason about dynamic environments~\citep{wang2024navformer, xu2025navrl}. These methods are flexible and adaptive in uncertain crowds, but safety and generalization remain difficult under unseen densities, agent behaviors, and body shapes. Safe RL methods add model-based lookahead or shielding to reduce violations~\citep{alshiekh2018safe}, but they do not directly provide explicit shape-aware safety terms. Consequently, few RL-based methods explicitly address arbitrarily shaped objects.

\textbf{Hybrid Methods.}
Hybrid methods aim to combine the adaptability of learning techniques with the structure of model-based collision avoidance~\citep{qin2023srl}. For example, reinforcement learning reciprocal velocity obstacle (\texttt{RL-RVO}) uses reciprocal velocity obstacle shaped rewards to guide distributed multi-robot policy learning~\citep{han2022reinforcement}. Another line couples learned decision making with MPC controller execution in social navigation~\citep{brito2021where}. Recent Deep Residual MPC (\texttt{DR-MPC}) blends MPC path tracking with model-free DRL for real-world navigation and uses out-of-distribution detection with a heuristic safety check to reduce unsafe exploration~\citep{han2025dr}. More generally, learning-based MPC can learn dynamics models, costs, constraints, or terminal value approximations, and can also use MPC as a safety layer around RL policies~\citep{hewing2020learning, zanon2020safe, xiao2025anycar, romero2025actor}. Unlike these methods, \texttt{SRL-MPC} does not learn the entire navigation policy or use optimization only as a shield for unsafe actions. Instead, RL adapts the HOCBF-MPC parameters from local GSFs, while the executed control is computed by the explicit shape-aware MPC optimization problem.

\section{Problem Statement}
\label{sec:problem}
Consider robot $i$ navigating in a local crowd of $N$ objects, including other robots, pedestrians, and static or dynamic obstacles. At each control cycle, robot $i$ plans a state sequence $\mathcal{S}_i=\{\mathbf{s}_{i,0},\ldots,\mathbf{s}_{i,T}\}$ and a control sequence $\mathcal{U}_i=\{\mathbf{u}_{i,0},\ldots,\mathbf{u}_{i,T-1}\}$ over an MPC horizon $T$, where $k$ denotes the prediction-step index. The state $\mathbf{s}_{i,k}$ contains the planar position $\mathbf{p}_{i,k}=[x_{i,k},y_{i,k}]^\top$ and heading $\theta_{i,k}$, and $\mathbf{u}_{i,k}$ is the control input. The feasible set $\mathcal{F}_i$ collects the kinematic model, input bounds, input-increment bounds, and initial condition. The world-frame occupied set of robot $i$ at step $k$ is denoted by $\mathbb{Z}_i(\mathbf{s}_{i,k})$, which is obtained from a body-frame convex set $\mathbb{C}_i$ by state transformation. Detailed kinematic and geometric definitions are given in Appendix~\ref{app:model_details}.

Let $\mathcal{N}_i$ be the neighbor set considered by robot $i$, and let the bar notation denote nominal or predicted neighbor quantities, e.g., $\bar{\mathbf{s}}_{j,k}$ and $\bar{\mathbf{p}}_{j,k}$ for neighbor $j$. The navigation objective is to follow a reference path while maintaining a desired control profile. Let $\mathcal{S}_i^{\mathrm{ref}}$ and $\mathcal{U}_i^{\mathrm{ref}}$ denote the reference state and control sequences. Following the path-tracking objective used in~\citep{han2025neupan}, the cost function is
\begin{equation}
\begin{aligned}
    C_i(\mathcal{S}_i,\mathcal{U}_i)
    &=
    w_p
    \lVert
    \mathbf{p}(\mathcal{S}_i)
    -
    \mathbf{p}(\mathcal{S}_i^{\mathrm{ref}})
    \rVert_2^2
    +
    w_\theta
    \lVert
    \theta(\mathcal{S}_i)
    -
    \theta(\mathcal{S}_i^{\mathrm{ref}})
    \rVert_2^2
    +
    w_u
    \lVert
    \mathcal{U}_i
    -
    \mathcal{U}_i^{\mathrm{ref}}
    \rVert_2^2 ,
\end{aligned}
\label{eq:cost}
\end{equation}
where $\mathbf{p}(\mathcal{S}_i)$ and $\theta(\mathcal{S}_i)$ denote the stacked positions and headings extracted from $\mathcal{S}_i$, respectively, and $w_p$, $w_\theta$, and $w_u$ balance position tracking, heading tracking, and control effort. This cost encourages the robot to progress toward the goal while maintaining the desired control profile.

Collision avoidance is imposed through the minimum distance between occupied sets. For each neighbor $j\in\mathcal{N}_i$ and step $k$, define
\begin{equation}
    D_{ij,k}
    =
    \min_{\mathbf{x}_i\in\mathbb{Z}_i(\mathbf{s}_{i,k}),\,
          \mathbf{x}_j\in\mathbb{Z}_j(\bar{\mathbf{s}}_{j,k})}
    \lVert \mathbf{x}_i-\mathbf{x}_j\rVert_2 ,
    \label{eq:collision_avoidance}
\end{equation}
where $\mathbf{x}_i$ and $\mathbf{x}_j$ are world-frame points on the two occupied sets, and $D_{ij,k}$ is the minimum Euclidean distance between robot $i$ and neighbor $j$. The required safety margin is denoted by $d_{\mathrm{safe}}$. The resulting local planning problem is
\begin{equation}
    \min_{\mathcal{S}_i,\mathcal{U}_i} C_i(\mathcal{S}_i,\mathcal{U}_i)
    \quad \mathrm{s.t.}\quad
    (\mathcal{S}_i,\mathcal{U}_i)\in\mathcal{F}_i,\;
    D_{ij,k}\ge d_{\mathrm{safe}},\;
    j\in\mathcal{N}_i,\; k=1,\ldots,T .
\label{eq:compact_primal_problem}
\end{equation}
where $(\mathcal{S}_i,\mathcal{U}_i)$ are the decision variables, $(\mathcal{S}_i,\mathcal{U}_i)\in\mathcal{F}_i$ enforces kinematic feasibility, and $D_{ij,k}\ge d_{\mathrm{safe}}$ enforces pairwise shape-aware separation along the horizon. This coupled problem is nonconvex and hard to solve directly. The next section introduces \texttt{SRL-MPC}, which separates geometry updates from local motion optimization and adapts the key MPC parameters online.
\begin{figure}[t]
    \centering
    \includegraphics[width=0.94\linewidth]{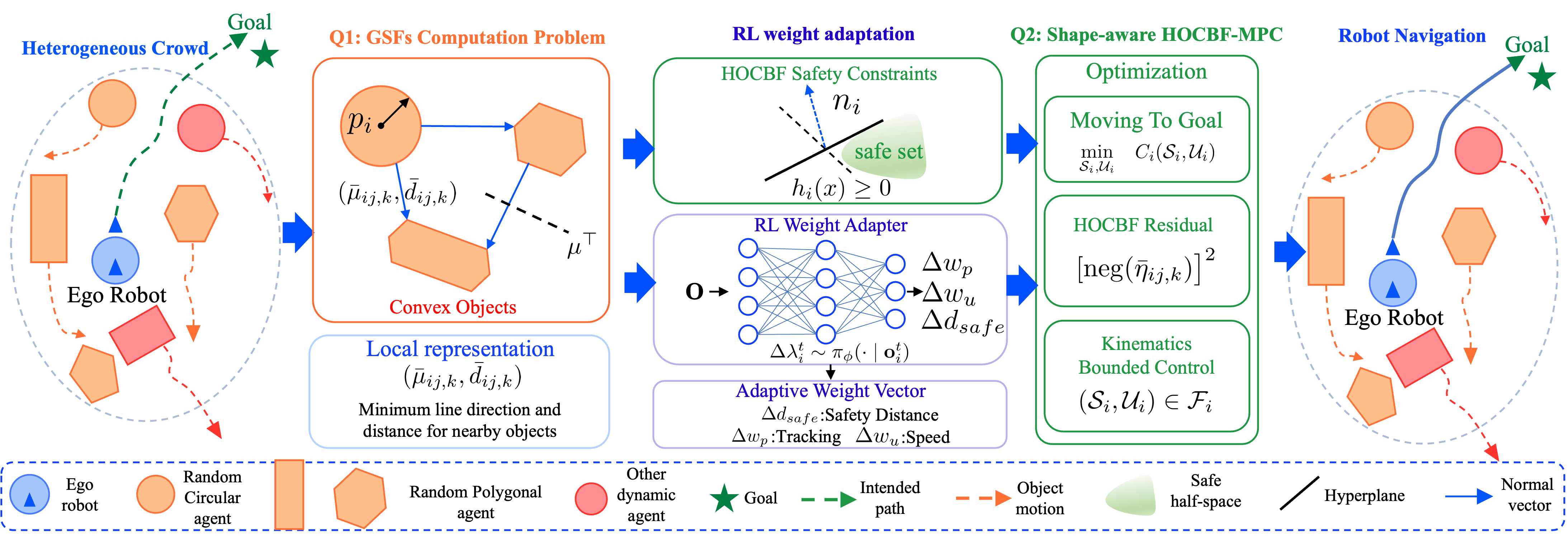}
    \caption{Overview of SRL-MPC. Key components: 1. A shape-aware HOCBF condition is formulated based on GSFs and works as a residual penalty in the optimization to improve safety; 2. the reinforcement learned policy adapts MPC parameters based on neighboring GSFs to handle dense scenarios.}
    \label{fig:method}
\end{figure}
\section{Method}
\label{sec:method}
The framework of the proposed method is illustrated in Figure~\ref{fig:method}. The method has two key components. First, it starts from the support function form of the pairwise minimum-distance problem and expresses the shape-aware safe set as degree-2 HOCBF constraints in the coupled primal problem. After problem decomposition, the GSFs are updated by a geometric solver, and the HOCBF condition is represented as a residual in the local MPC problem. Second, the reinforcement learning framework maps GSFs to MPC parameter updates, adapting tracking weights, desired speed weights, and the HOCBF safety distance to autonomously balance task completion efficiency and safety.

\subsection{Shape-Aware HOCBF Constraint}
Few methods consider shape-aware HOCBF constraints because the exact conic constraints grow with the number of shape edges. Here we introduce the support function transformation~\citep{boyd2004convex} to formulate shape-aware HOCBF constraints based on fixed-dimensional GSFs.
For a pair $(i,j)$ and prediction step $k$, let $\bar{\mathbf{s}}_{j,k}$ be the nominal state of neighbor $j$. Under the strong-duality conditions used in~\citep{zhang2020optimization, han2023rda}, the minimum-distance computation in~\eqref{eq:collision_avoidance}, equivalently the body-frame program in~\eqref{eq:app_body_frame_distance}, can be rewritten in dual form. To avoid the conic constraints, we further transform it into a support function form by:
\begin{equation}
\begin{aligned}
    D_{ij,k}
    =
    \max_{\lVert \mu \rVert_{*} \le 1}
    \Phi_{ij,k}(\mu),\quad
    \Phi_{ij,k}(\mu)
    =
    \mu^\top(\mathbf{p}_{i,k} - \bar{\mathbf{p}}_{j,k})
    - \sigma_{\mathbb{C}_i}(-\mathbf{R}_{i,k}^\top\mu)
    - \sigma_{\mathbb{C}_j}(\bar{\mathbf{R}}_{j,k}^\top\mu),
\end{aligned}
\label{eq:dual_distance}
\end{equation}
where $\mu^\top$ represents the separating hyperplane between two convex occupied sets, $\mu$ is the minimum-distance direction, and $\lVert\cdot\rVert_*$ denotes the dual norm. We define the geometric separation feature as $\mathrm{GSF}_{ij,k}\triangleq(\mu_{ij,k},D_{ij,k})$, where $\mu_{ij,k}\in\arg\max_{\lVert\mu\rVert_*\le1}\Phi_{ij,k}(\mu)$ and $D_{ij,k}$ is the corresponding minimum distance.
For a generic body-frame occupied set $\mathbb{C}\subset\mathbb{R}^2$ and a query direction $\boldsymbol{\xi}\in\mathbb{R}^2$, the support function $\sigma_{\mathbb{C}}(\boldsymbol{\xi})=\sup\{\,\boldsymbol{\xi}^{\top}\mathbf{z}\mid \mathbf{z}\in\mathbb{C}\,\}$ gives the maximum projection of $\mathbb{C}$ along $\boldsymbol{\xi}$. Equation~\eqref{eq:dual_distance} therefore expresses the shape-aware minimum distance through the maximum projection induced by the separating hyperplane $\mu^\top$. Subtracting the required safety margin $d_{\mathrm{safe}}$ gives:
$h_{ij,k}(\mu)
    =
    \Phi_{ij,k}(\mu)-d_{\mathrm{safe}} .$
For a selected $\mu_{ij,k}$, the collision avoidance condition at step $k$ is written as $h_{ij,k}(\mu_{ij,k})\ge 0$. Thus, the shape-aware safe set at prediction step $k$ is
\begin{equation}
\begin{aligned}
    \mathcal C_{ij,k}
    &=
    \{(\mathbf{s}_{i,k},\bar{\mathbf{s}}_{j,k}) \mid H_{ij,k}\ge 0\},
    \qquad
    H_{ij,k}
    \triangleq
    h_{ij,k}(\mu_{ij,k}) .
\end{aligned}
\label{eq:pairwise_safety_set}
\end{equation}
A first-order discrete control barrier function (CBF) only imposes a one-step condition on the barrier value, which is weak for dense dynamic interactions with a high-speed profile. We therefore use a degree-2 discrete HOCBF~\citep{xiao2022high}. For $\alpha_1,\alpha_2\in(0,1]$, the degree-2 HOCBF left-hand side is denoted as
\begin{equation}
\begin{aligned}
    \eta_{ij,k}
    &=
    H_{ij,k+2}
    -
    (2-\alpha_1-\alpha_2)H_{ij,k+1}
    +
    (1-\alpha_1)(1-\alpha_2)H_{ij,k}.
\end{aligned}
\label{eq:hocbf_lhs}
\end{equation}
In the primal problem, the HOCBF condition is imposed as the hard safety constraint $\eta_{ij,k}\ge 0$.
With $\alpha_1=\alpha_2=\gamma$ and $\gamma\in(0,1]$, this constraint becomes
\begin{equation}
    H_{ij,k+2}
    -
    2(1-\gamma)H_{ij,k+1}
    +
    (1-\gamma)^2 H_{ij,k}
    \ge 0,
    \qquad
    j\in\mathcal{N}_i^{\mathrm{cbf}},
    \; k=1,\ldots,H_c .
    \label{eq:hocbf_gamma}
\end{equation}
where $\mathcal{N}_i^{\mathrm{cbf}}\subseteq\mathcal{N}_i$ is the selected nearest-neighbor subset used for HOCBF terms, and $H_c\le T-2$ is the HOCBF horizon.

\textbf{Problem Decomposition:}
The minimum-distance line direction $\mu$ and the MPC trajectory $\mathbf{p}$ are coupled in the HOCBF constraint, leading to a bi-convex optimization problem. Following the block decomposition idea used in~\citep{han2023rda, han2025neupan}, \texttt{SRL-MPC} separates the coupled problem into a GSF update subproblem $Q_1$ and a local HOCBF-MPC subproblem $Q_2$. The first block fixes nominal trajectories and computes nominal GSFs, while the second block fixes $\mu$ and computes the action sequence. After this decomposition, the HOCBF condition is handled through a soft residual penalty in $Q_2$. By iteratively solving $Q_1$ and $Q_2$, the geometric features and local trajectory are updated consistently, providing stronger optimization guidance than a static distance-margin penalty.

\textbf{$Q_1$: GSF computation subproblem.}
Given nominal trajectories $\bar{\mathcal{S}}_i$ and $\bar{\mathcal{S}}_j$, $Q_1$ computes the nominal GSFs between two transformed convex occupied sets. Instead of solving the optimization program, which is computationally expensive, we use a geometry-based shortest-line computation implemented by the Geometry Engine Open Source (GEOS) library~\citep{geos}, which is efficient for geometries:
\begin{equation}
    \overline{\mathrm{GSF}}_{ij,k}
    \triangleq
    (\bar{\mu}_{ij,k},\bar{d}_{ij,k})
    =
    \operatorname{ShortestLine}\big(\mathbb{Z}_i(\bar{\mathbf{s}}_{i,k}),\mathbb{Z}_j(\bar{\mathbf{s}}_{j,k})\big),
    \label{eq:shortest_line}
\end{equation}
where $\operatorname{ShortestLine}(\cdot)$ denotes the geometric shortest-line algorithm that returns the unit direction from the neighbor set to the ego set and the corresponding minimum distance. Specifically, for a circular set, $\bar{\mu}_{ij,k}$ can be obtained in closed form from the normalized center difference and the radius.
Neighbors considered in the local MPC problem are ranked by the minimum distance over a short distance horizon, and only the closest $K$ neighbors are retained in $\mathcal{N}_i$, where $K$ is a user-specified neighbor budget.

\textbf{$Q_2$: local HOCBF-MPC subproblem.}
Given the GSFs produced by $Q_1$, the local HOCBF-MPC subproblem $Q_2$ is formulated as
\begin{equation}
    \begin{aligned}
    Q_2:\quad
    \min_{\mathcal{S}_i,\mathcal{U}_i}\quad
    & C_i(\mathcal{S}_i,\mathcal{U}_i)
    +\frac{\rho_{\mathrm{obs}}}{2}
    \sum_{j\in\mathcal{N}_i^{\mathrm{cbf}}}\sum_{k=1}^{H_c}
    [\operatorname{neg}(\bar{\eta}_{ij,k})]^2,\\
    \mathrm{s.t.}\quad
    &(\mathcal{S}_i,\mathcal{U}_i)\in\mathcal{F}_i .
    \end{aligned}
    \label{eq:cbf_optimization_problem}
\end{equation}
where $\rho_{\mathrm{obs}}>0$ is the penalty weight for HOCBF violation and is selected as a large value (e.g., 100). $\operatorname{neg}(x)=\max(0,-x)$ is the negative-part operator, and $\bar{\eta}_{ij,k}$ is the fixed-geometry counterpart of~\eqref{eq:hocbf_lhs} after substituting the GSFs produced by $Q_1$.
The penalty is zero when the decomposed high-order condition is satisfied and positive only when the predicted trajectory violates it. The soft penalty form avoids infeasibility that can occur when high-order conditions are imposed as hard local constraints in dense crowds, while keeping the local feasible set $\mathcal{F}_i$ unchanged. With the geometric terms fixed by $Q_1$ and the kinematics represented by affine linearization, $Q_2$ is a convex local HOCBF-MPC subproblem.

At each control cycle, the solver alternates these two subproblems for a small number of iterations. Nominal trajectories are first propagated from the current states. Then, for each robot, $Q_1$ updates the GSFs for the selected neighbors, $Q_2$ solves the local HOCBF-MPC problem, and the resulting trajectory becomes the nominal trajectory for the next iteration. The first control in the optimized sequence is applied to the controlled object.

\subsection{Reinforcement Learned MPC}
\begin{figure}[t]
    \centering
    \includegraphics[width=0.95\linewidth]{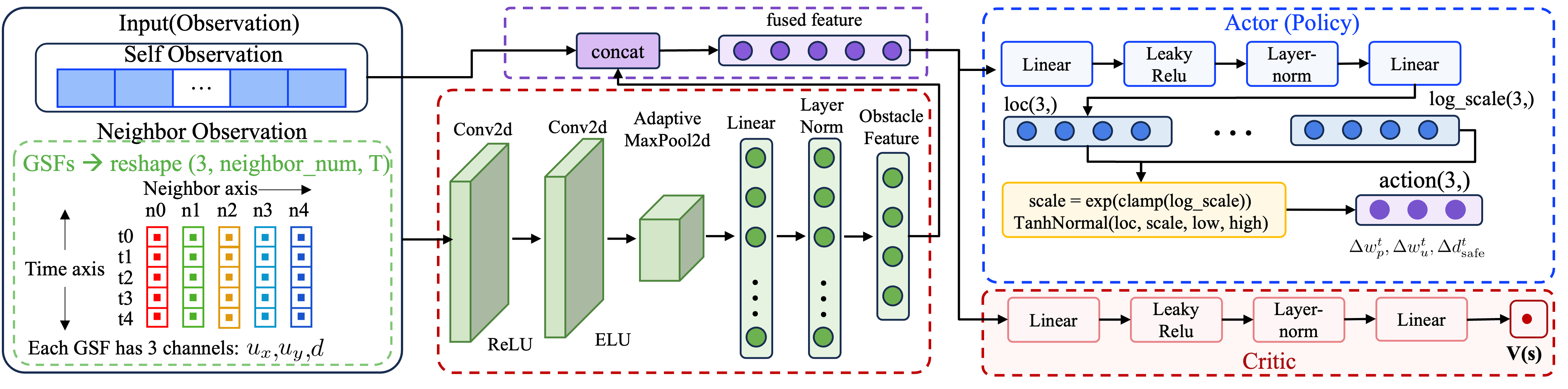}
    \caption{Neural network architecture. The GSFs are encoded by CNNs and then concatenated with the self-observation to form a fused feature, which is shared by the actor and critic.}
    \label{fig:neural_network}
\end{figure}
The solution of $Q_2$ depends strongly on the parameters in the MPC cost and safety terms, including $w_p$, $w_\theta$, $w_u$, and $d_{\mathrm{safe}}$. These weights and safety parameters determine how aggressively or conservatively a robot tracks the reference path (i.e., $w_p$), how strongly it penalizes control effort (i.e., $w_u$), and how much safety margin it requests from nearby agents (i.e., $d_{\mathrm{safe}}$). These parameters usually require repeated manual tuning for each scenario. For example, in open space, larger $w_u$ and $w_p$ can guide the robot to track the reference path toward the goal quickly, while in dense scenarios, smaller $w_p$ or $d_{\mathrm{safe}}$ may be needed to avoid collisions or getting stuck. This work proposes an RL-based parameter adaptation method to automatically tune these parameters, as illustrated in Figure~\ref{fig:neural_network}, while retaining the explicit geometric and kinematic structure and generalizability of MPC. The learned policy observes local GSFs and current parameter values, outputs bounded parameter increments, and leaves control synthesis to MPC. Because the GSF tensor stacks the closest neighbors over the horizon, the policy can adapt the MPC behavior to local crowd density and shape geometry.

\textbf{Action space:}
The adaptive parameters are $w_p$, $w_u$, and $d_{\mathrm{safe}}$, while $w_\theta$ is fixed to a small value, e.g., $0.01$, to encourage robots to turn during collision avoidance. To make the parameters change smoothly over time, the policy outputs parameter increments rather than absolute parameter values. Thus, for robot $i$ at time step $t$, the policy output is $\Delta\lambda_i^t
=
\begin{bmatrix}
    \Delta w_{p,i}^t & \Delta w_{u,i}^t & \Delta d_{\mathrm{safe},i}^t
\end{bmatrix}^{\top}.$ The parameter vector is updated as
\begin{equation}
    \lambda_i^{t+1}
    =
    \lambda_i^t+\Delta\lambda_i^t
    \in \Lambda,\quad
    \Lambda=[0.01,1.0]\times[0.1,10.0]\times[0.2,1.0].
    \label{eq:parameter_update}
\end{equation}
The increment is bounded by $\Delta\lambda_i^t\in[-0.5,0.5]\times[-0.5,0.5]\times[-0.1,0.1]$, where the third component is measured in meters. The absolute parameter values are clipped to the admissible set $\Lambda$, where $w_p\in[0.01,1.0]$, $w_u\in[0.1,10.0]$, and $d_{\mathrm{safe}}\in[0.2,1.0]$. A larger range is assigned to $w_u$ to allow the policy to adjust the control effort strongly in dense scenarios.

\textbf{Observation space:}
The robot observation $\mathbf{o}_i^t$ has two parts: the self-observation $\mathbf{o}_{i,\mathrm{self}}^t$ and the neighbor observation $\mathbf{o}_{i,\mathrm{neighbor}}^t$:
\begin{equation}
\begin{gathered}
    \mathbf{o}_{i,\mathrm{self}}^t
    =
    \begin{bmatrix}
    v_i^t &
    w_{p,i}^t &
    w_{u,i}^t &
    d_{\mathrm{safe},i}^t
    \end{bmatrix}^{\top},
    \quad
    \mathbf{o}_{ij,k,\mathrm{neighbor}}^t
    =
    \begin{bmatrix}
        \bar{\mu}_{ij,k,x}^t &
        \bar{\mu}_{ij,k,y}^t &
        \bar{d}_{ij,k}^t
    \end{bmatrix}^{\top}.
\end{gathered}
\label{eq:rl_observation}
\end{equation}
Here $\mathbf{o}_{ij,k,\mathrm{neighbor}}^t$ stores GSFs, i.e., $(\bar{\mu}_{ij,k}^t, \bar{d}_{ij,k}^t)$ for neighbor $j$ at horizon step $k$. Thus, $\mathbf{o}_{i,\mathrm{neighbor}}^t\in\mathbb{R}^{3\times K\times T}$ stacks these geometric features for the $K$ closest neighbors over the MPC horizon. The neighbors are selected by the minimum predicted distance over a short distance horizon, and the tensor is zero-padded when fewer than $K$ neighbors are active. The self-observation includes the current MPC parameters $w_{p,i}^t$, $w_{u,i}^t$, $d_{\mathrm{safe},i}^t$, and the forward speed $v_i^t$. Compared with general RL-based approaches~\citep{chen2019crowd, han2022reinforcement, xu2025navrl}, this observation space is compact and does not require explicit pose or goal-direction features. This compactness is possible because goal tracking and collision avoidance are handled by HOCBF-MPC; RL only learns how to adapt the MPC parameters to changing obstacle distributions.

\textbf{Neural network design.}
The neural network architecture is shown in Figure~\ref{fig:neural_network}. The neighbor observation $\mathbf{o}_{i,\mathrm{neighbor}}^t$ is treated as a three-channel image over the neighbor-horizon grid. The encoder applies two padded $3\times3$ convolutional layers, using a rectified linear unit (ReLU) after the first convolution and an exponential linear unit (ELU) after the second. This gives each output cell access to the full neighbor-horizon grid before pooling. An adaptive max-pooling layer keeps the strongest local interaction response, and a linear projection followed by layer normalization produces a fixed-size geometric feature.

The encoded geometric feature is concatenated with the 4-dimensional self-observation to form a shared feature. This shared encoder is used by both the actor and the critic. The actor outputs a 6-dimensional vector that is split into a 3-dimensional location vector and a 3-dimensional log-scale vector for the parameter increment $\Delta\lambda_i^t$. The resulting squashed Gaussian distribution samples bounded parameter increments.
The critic uses the same shared encoder and a parallel linear head with LeakyReLU and layer normalization to output the scalar value estimate for proximal policy optimization (PPO) advantage computation. Layer normalization is used because the observation contains raw physical units, including velocity, MPC weights, and safety distance.

\textbf{Reward design:}
The reward function is designed to align policy learning with task completion and safety. It uses a one-shot arrival bonus, a collision penalty, a small per-step time penalty, and a safety log-barrier penalty:
\begin{equation}
    \begin{aligned}
    r_i^t =
    r_{\mathrm{arr}}\mathbf{1}\{i\ \mathrm{arrives}\}
    -r_{\mathrm{col}}\mathbf{1}\{i\ \mathrm{collides}\}
    -r_{\mathrm{safe}}\ell_{\mathrm{safe},i}^t
    -r_{\mathrm{step}},\\
    \ell_{\mathrm{safe},i}^t =
    \mathbf{1}\{d_{i,\mathrm{near}}^t<d_{\mathrm{safe},i}^t\}
    \operatorname{clip}_{[0,1]}
    \left(
    \frac{\log\!\left(d_{\mathrm{safe},i}^t/\max(d_{i,\mathrm{near}}^t,\epsilon_d)\right)}
    {\log\!\left(d_{\mathrm{safe}}^{\max}/d_{\mathrm{safe}}^{\min}\right)}
    \right),
    \end{aligned}
    \label{eq:rl_reward}
\end{equation}
Here $\mathbf{1}\{\cdot\}$ is a binary indicator, $d_{i,\mathrm{near}}^t$ is the current minimum distance from robot $i$ to its nearest neighbor, $d_{\mathrm{safe},i}^t$ is the policy-adapted safety distance, and $\epsilon_d$ is a small numerical floor. The safety term is active only when the current minimum distance is smaller than the selected safety distance. The arrival bonus is assigned once when the robot reaches its goal and then parks at the goal, while the collision penalty is assigned when a collision is detected. The collision penalty is larger than the arrival bonus, encouraging the policy to choose MPC parameters that reach the goal without relying on unsafe behaviors. The numerical reward constants are listed in Table~\ref{tab:rl_parameters} of Appendix~\ref{app:experiment_parameters}.

\begin{wrapfigure}[10]{r}{0.35\textwidth}
    \centering
    \includegraphics[width=0.465\linewidth]{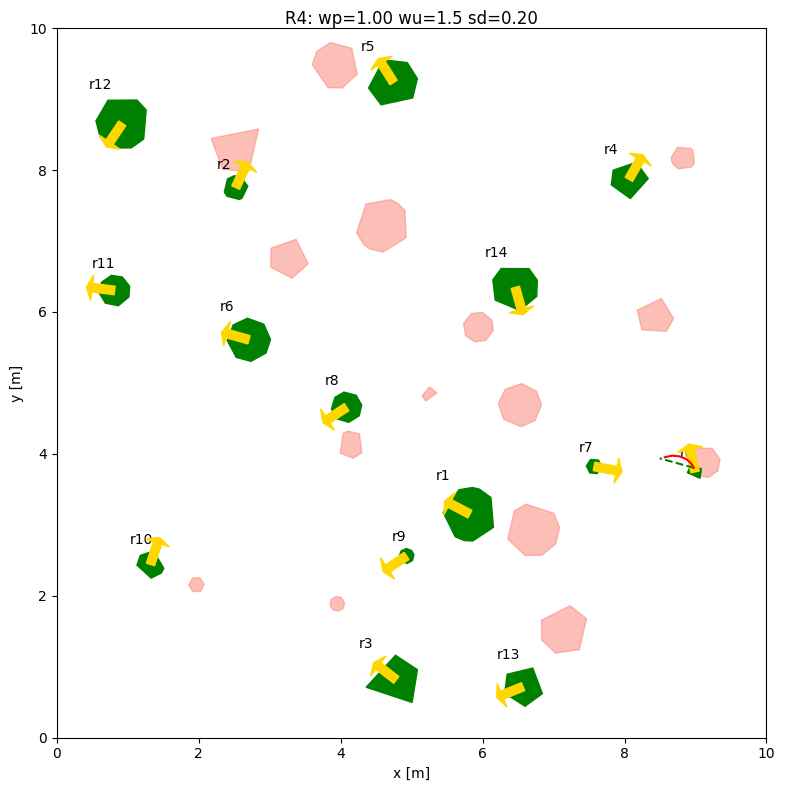}
    \hspace{1pt}
    \includegraphics[width=0.465\linewidth]{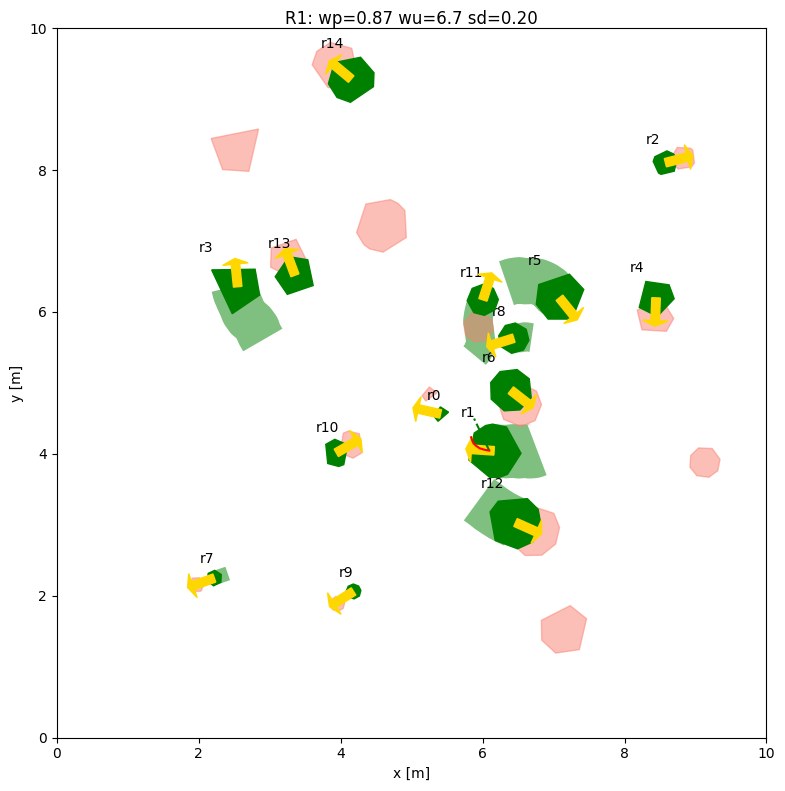}
    \caption{Training scenario examples. Each episode randomly samples start positions, goal positions, and convex polygon footprints; arrived robots park as static obstacles.}
    \label{fig:training_scenario}
\end{wrapfigure}

\textbf{Training:}
The policy is trained with PPO~\citep{schulman2017proximal} using decentralized execution and shared parameters. All robots share the same network parameters, and each robot reads the observation in~\eqref{eq:rl_observation} to generate its own MPC parameter increment. The training scenario is generated by IR-SIM~\citep{han2026ir} and shown in Figure~\ref{fig:training_scenario}: each episode samples $15$ robots in a $10\,\mathrm{m}\times10\,\mathrm{m}$ workspace with random start-goal pairs and convex polygon robot footprints. Each robot stops at its goal after arrival and then acts as a static obstacle for the other robots. A robot is reset when it collides or times out. Compared with other RL-based approaches trained with circular robots, sparse dynamic obstacles, or a fixed workspace, the training scenario is challenging because it combines randomized geometry, dense crowds, and both static and dynamic obstacles. The training process takes about $4$ hours to achieve stable performance.

\section{Experiments}
\label{sec:experiments}
\begin{figure}[t]
    \centering
    \begin{minipage}{0.158\linewidth}
        \centering
        \includegraphics[width=0.95\linewidth]{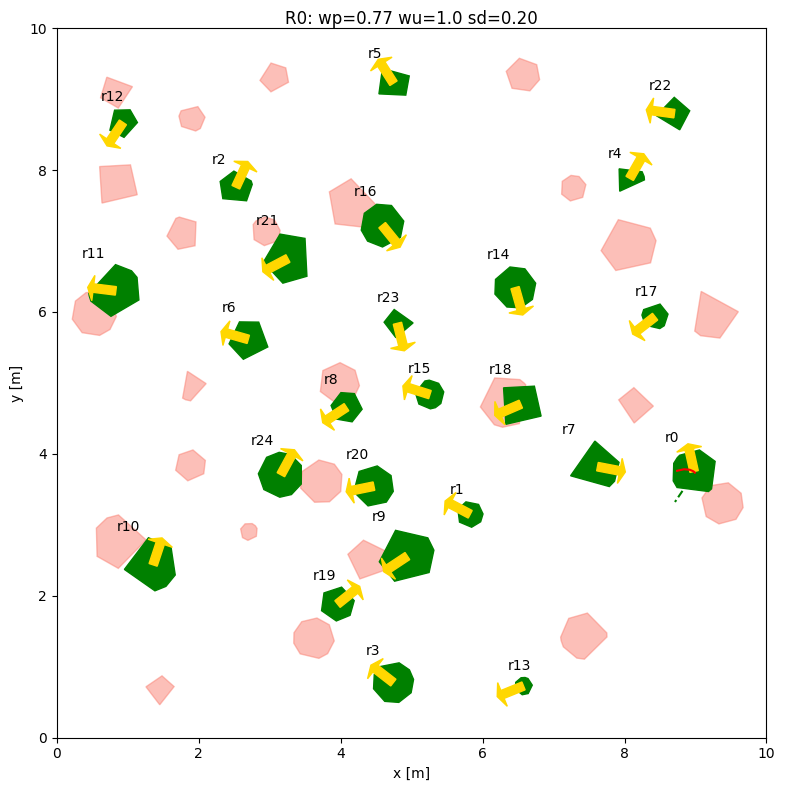}
        \par{\scriptsize Step 01}
    \end{minipage}
    \hfill
    \begin{minipage}{0.158\linewidth}
        \centering
        \includegraphics[width=0.95\linewidth]{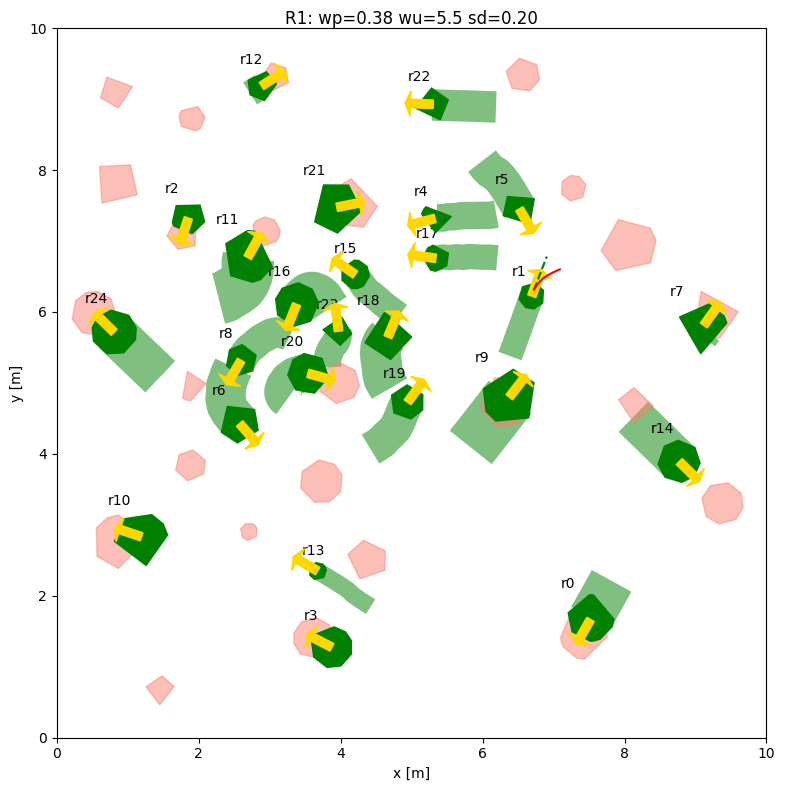}
        \par{\scriptsize Step 35}
    \end{minipage}
    \hfill
    \begin{minipage}{0.158\linewidth}
        \centering
        \includegraphics[width=0.95\linewidth]{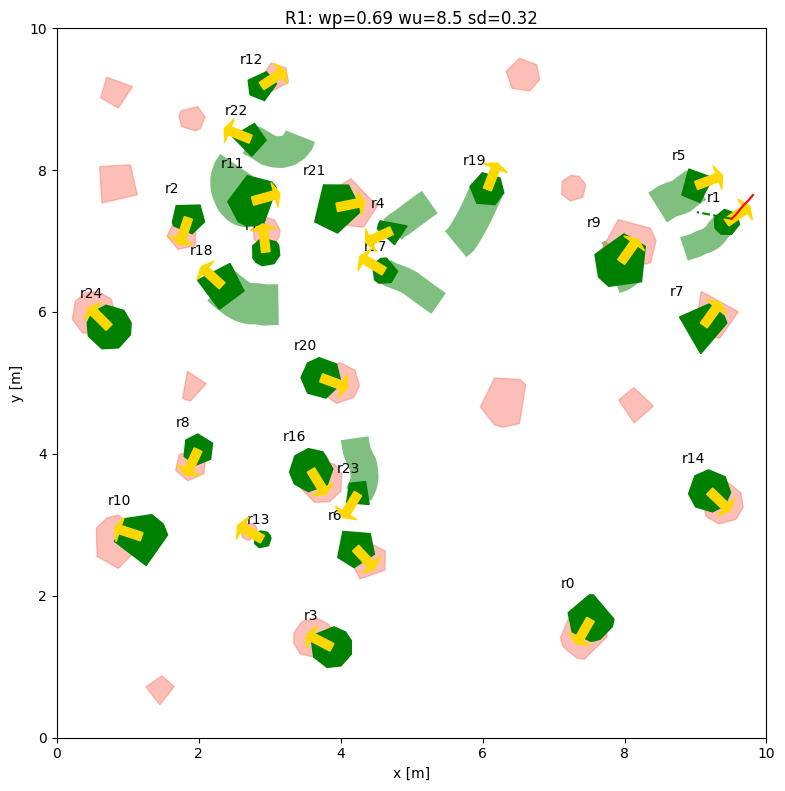}
        \par{\scriptsize Step 70}
    \end{minipage}
    \hfill
    \begin{minipage}{0.158\linewidth}
        \centering
        \includegraphics[width=0.95\linewidth]{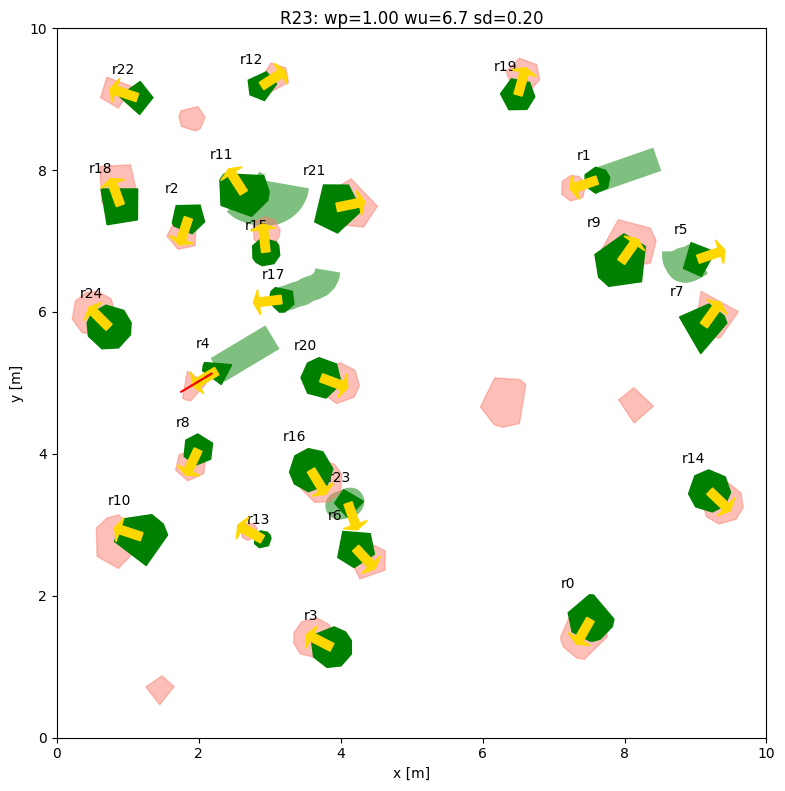}
        \par{\scriptsize Step 109}
    \end{minipage}
    \hfill
    \begin{minipage}{0.158\linewidth}
        \centering
        \includegraphics[width=0.95\linewidth]{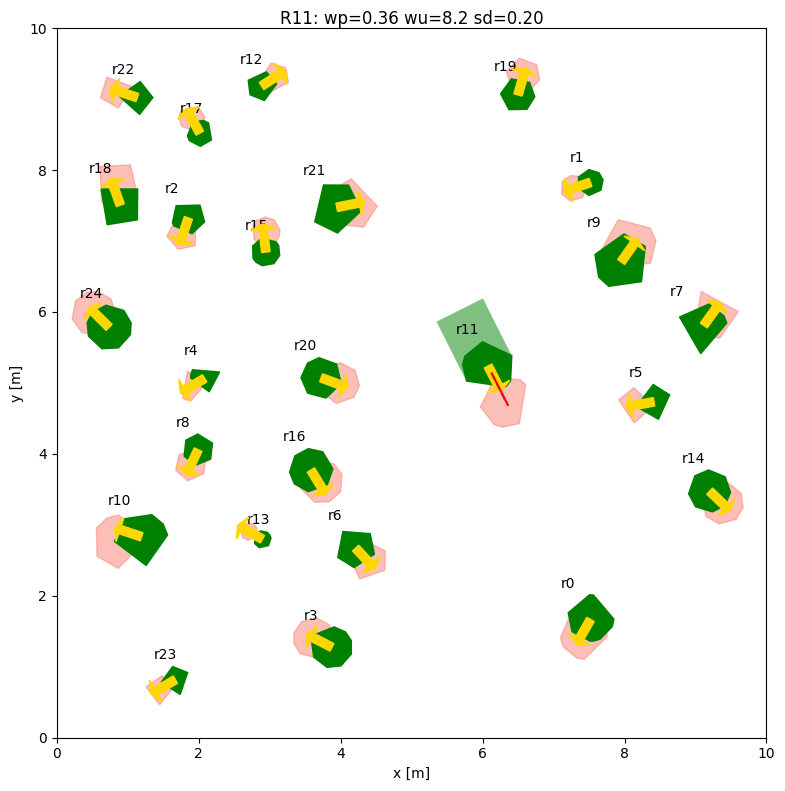}
        \par{\scriptsize Step 188}
    \end{minipage}
    \par{\small{(a)}}
    \vspace{1pt}

    \begin{minipage}{0.158\linewidth}
        \centering
        \includegraphics[width=0.95\linewidth]{exp_eval_188}
        \par{\scriptsize Random}
    \end{minipage}
    \hfill
    \begin{minipage}{0.158\linewidth}
        \centering
        \includegraphics[width=0.95\linewidth]{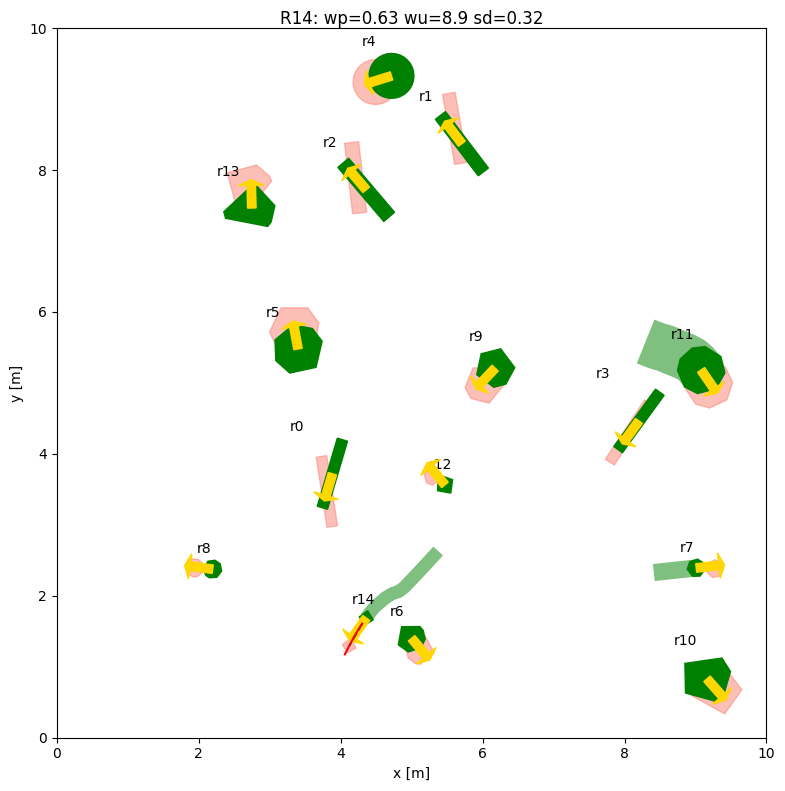}
        \par{\scriptsize Cross}
    \end{minipage}
    \hfill
    \begin{minipage}{0.158\linewidth}
        \centering
        \includegraphics[width=0.95\linewidth]{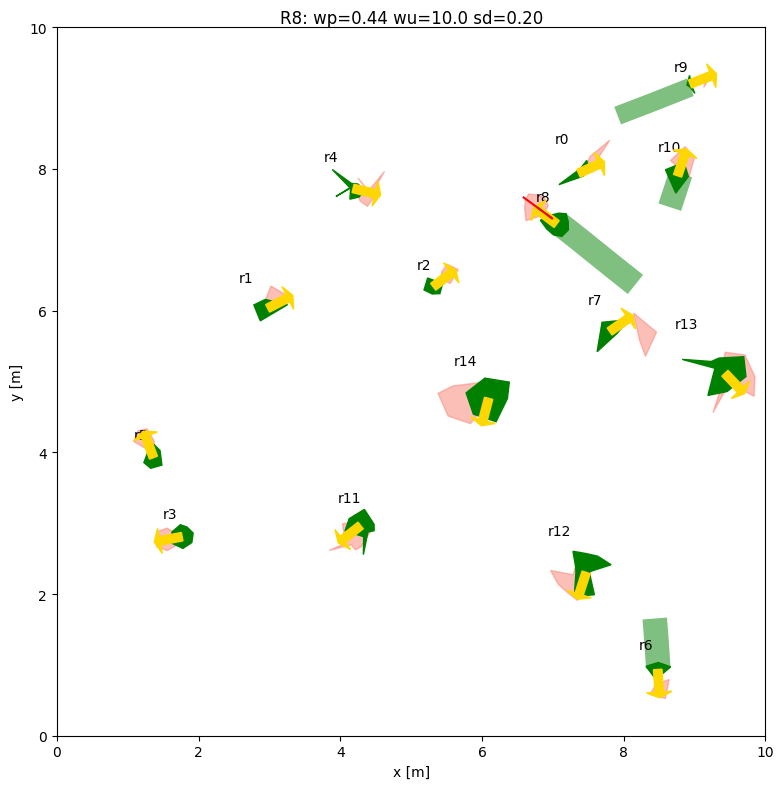}
        \par{\scriptsize Nonconvex}
    \end{minipage}
    \hfill
    \begin{minipage}{0.158\linewidth}
        \centering
        \includegraphics[width=0.95\linewidth]{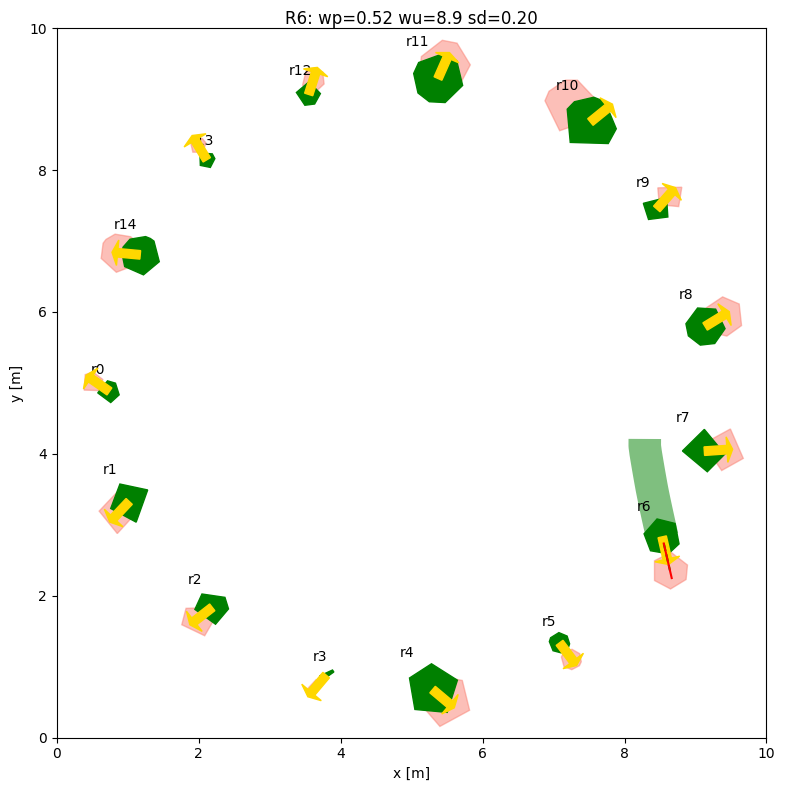}
        \par{\scriptsize Circular}
    \end{minipage}
    \hfill
    \begin{minipage}{0.158\linewidth}
        \centering
        \includegraphics[width=0.95\linewidth]{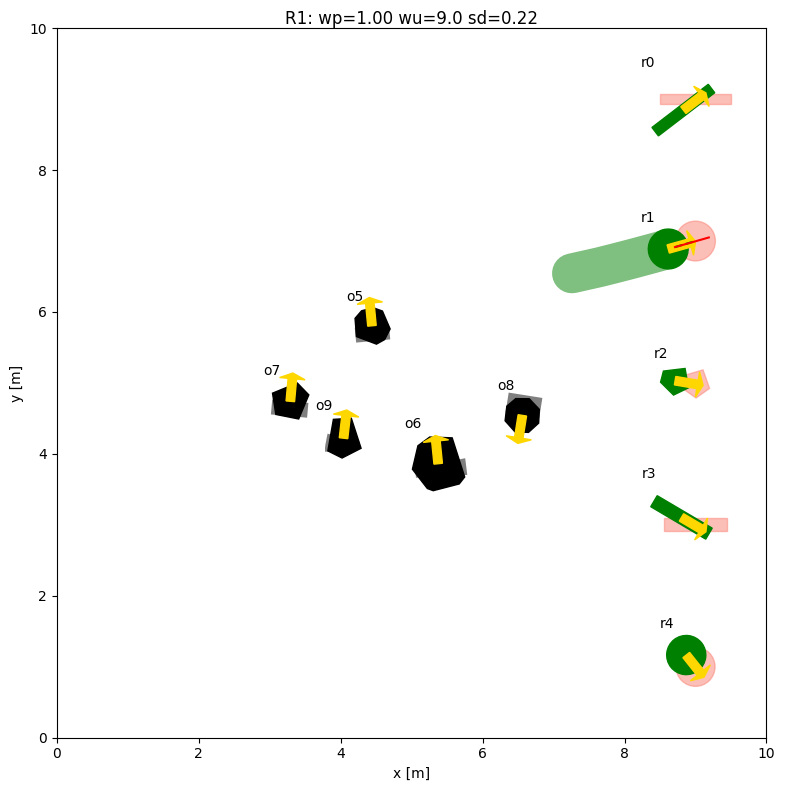}
        \par{\scriptsize Traffic}
    \end{minipage}
    \par{\small{(b)}}
    \caption{(a) Evaluation rollout with $25$ randomly generated polygonal robots. (b) Random-polygon and out-of-distribution scenarios, including cross geometry, nonconvex union, circular polygon, and through traffic. Full rollouts are shown in Appendix~\ref{app:additional_scenarios}.}
    \label{fig:evaluation_scenario}
\end{figure}

\textbf{Evaluation:}
We evaluate \texttt{SRL-MPC} in the same IR-SIM~\citep{han2026ir} randomized scenario family used for training, but with held-out random seeds, so no evaluation episode is reused during policy training. To test scalability and density robustness within this randomized shape-aware setting, the robot count is swept over $N\in\{10,15,20,25\}$ to represent different crowd densities, while using the same model trained only in 15-robot scenarios.
Figure~\ref{fig:evaluation_scenario} (a) shows a rollout with $25$ randomly generated polygonal robots; Figure~\ref{fig:evaluation_scenario} (b) shows examples of out-of-distribution (OOD) scenarios. The OOD scenarios include cross geometry, nonconvex-union footprints, circular exchange, and through-traffic interactions, covering different geometric and crowd-flow patterns beyond the main random-polygon density sweep. Guided by \texttt{SRL-MPC}, each robot adapts its parameters to different neighbor geometries and reaches its goal successfully. Even in congested situations caused by parked robots, the ego robot adjusts its parameters to detour around the obstacles.

The main density-sweep evaluations use $100$ episodes per robot count, a maximum episode length of $500$ control steps, and the same random seeds across compared methods. An episode is counted as successful only when all robots reach their goals without collision or timeout. Navigation time and path length are averaged over arrived robots, while speed is averaged over active robot frames. All $\pm$ values denote standard deviations over the corresponding evaluated samples. The main simulation experiments are conducted on a MacBook Pro with an Apple M4 Pro CPU. The detailed controller, training, and evaluation parameters are summarized in Tables~\ref{tab:controller_parameters}--\ref{tab:evaluation_parameters} of Appendix~\ref{app:experiment_parameters}.

\begin{figure}[t]
    \centering
    \includegraphics[width=0.92\linewidth]{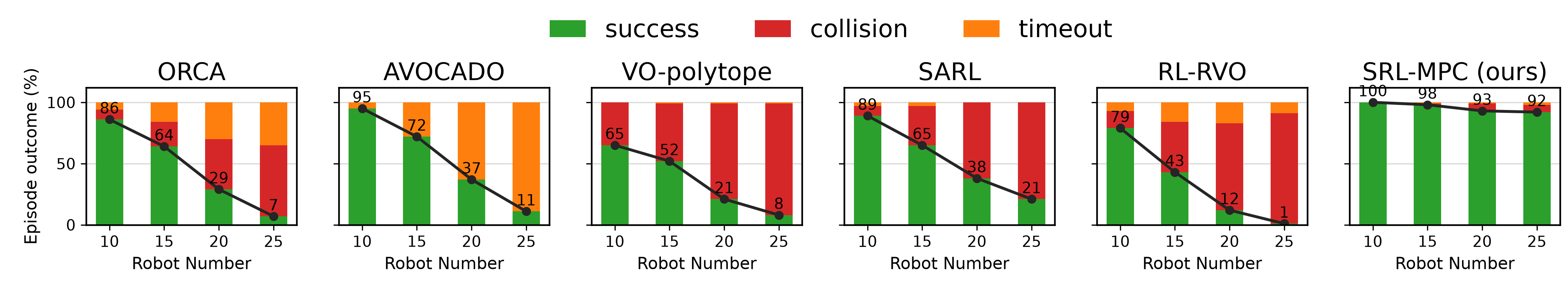}
    \caption{Episode outcomes for baseline methods in the random convex polygon scenario.}
    \label{fig:baseline_comparison}
\end{figure}

\begin{wrapfigure}[10]{r}{0.3\textwidth}
    \centering
    \includegraphics[width=\linewidth]{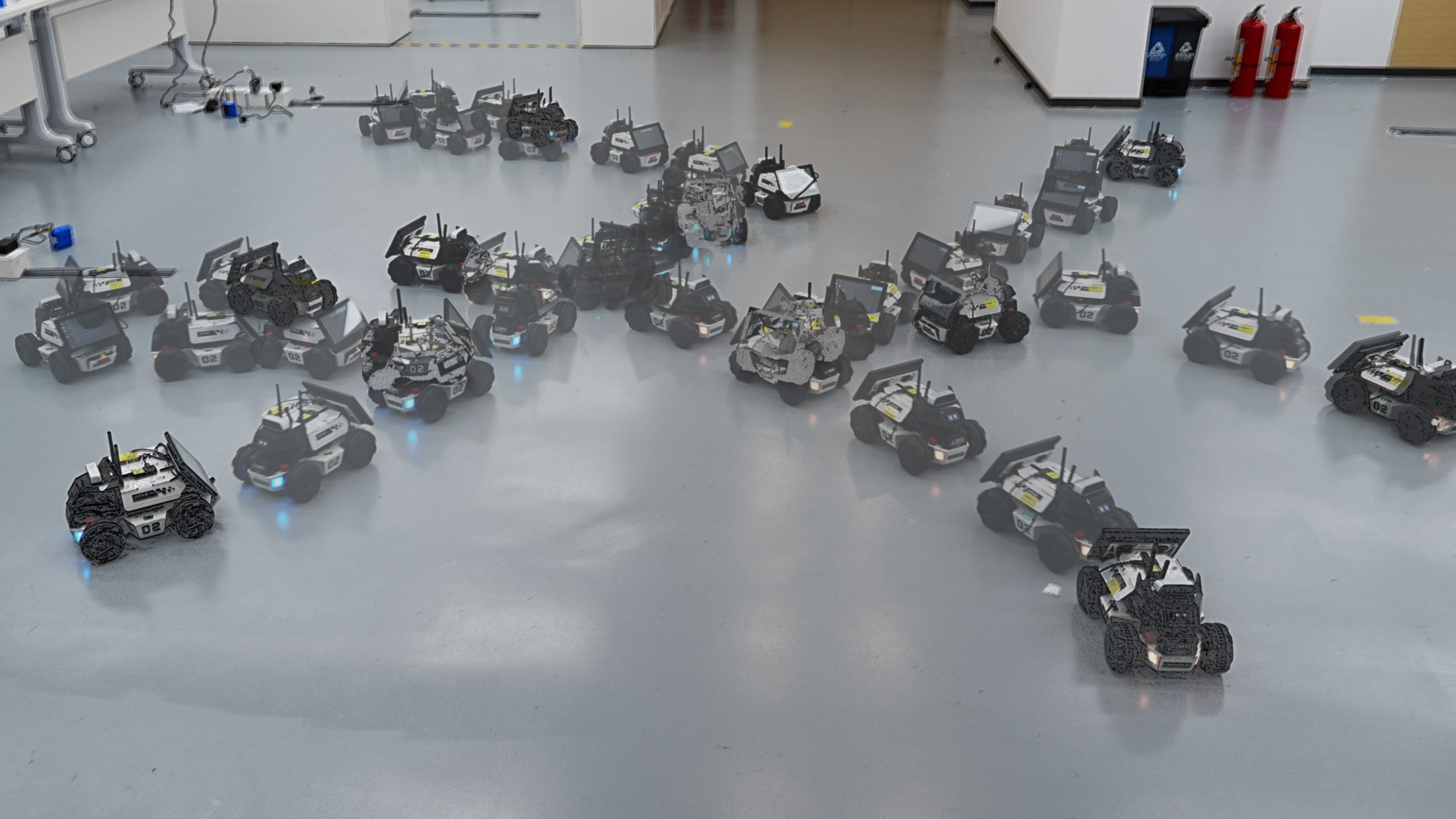}
    \caption{Real-world platform deployment.}
    \label{fig:real_world_experiment}
\end{wrapfigure}

\textbf{Comparison with Baselines:}
We compare \texttt{SRL-MPC} with five representative baselines: \texttt{ORCA}, a classic reciprocal velocity obstacle method~\citep{van2011reciprocal}; \texttt{AVOCADO}, an adaptive optimal collision avoidance method with opinion dynamics~\citep{martinez2025avocado}; \texttt{VO-polytope}, a polygon-precise velocity obstacle method~\citep{huang2023velocity}; \texttt{SARL}, a socially aware reinforcement learning policy that uses attention to encode crowd interactions~\citep{chen2019crowd}; and \texttt{RL-RVO}, a pretrained reinforcement learning policy based on reciprocal velocity obstacle features~\citep{han2022reinforcement}. These baselines cover reactive, geometry-aware, learning-based, and adaptive collision avoidance methods commonly used in crowd navigation. Figure~\ref{fig:real_world_experiment} shows the real-world experiment setup.
\begin{table}[t!]
\caption{Baseline comparison in the random-polygon setting}
\label{tab:baseline_results}
\centering
\tiny
\setlength{\tabcolsep}{2.0pt}
\renewcommand{\arraystretch}{0.84}
\global\setbox\baselineTableBox=\hbox{
\begin{tabular}{lc|ccc|ccc}
\specialrule{1pt}{1pt}{0pt}
\multicolumn{1}{c}{Robots} & \multicolumn{1}{c|}{Method} &
Success$\uparrow$ & Collision$\downarrow$ & Timeout$\downarrow$ &
Time (s)$\downarrow$ & Speed (m/s)$\uparrow$ & Path (m)$\downarrow$ \\
\midrule
\multirow{6}{*}{10} & \texttt{ORCA}~\citep{van2011reciprocal} & 86.0 & 8.0 & 6.0 & $5.09\pm2.58$ & $0.878\pm0.137$ & $4.14\pm2.25$ \\
& \texttt{AVOCADO}~\citep{martinez2025avocado} & 95.0 & 0.0 & 5.0 & $5.61\pm2.63$ & $0.785\pm0.137$ & $4.58\pm2.28$ \\
& \texttt{VO-polytope}~\citep{huang2023velocity} & 65.0 & 35.0 & 0.0 & $7.46\pm3.91$ & $0.536\pm0.067$ & $4.14\pm2.25$ \\
& \texttt{SARL}~\citep{chen2019crowd} & 89.0 & 8.0 & 3.0 & $8.90\pm5.24$ & $0.664\pm0.164$ & $5.78\pm3.11$ \\
& \texttt{RL-RVO}~\citep{han2022reinforcement} & 79.0 & 14.0 & 7.0 & $6.99\pm4.36$ & $0.780\pm0.136$ & $5.42\pm2.96$ \\
\rowcolor[gray]{0.95}
& \texttt{SRL-MPC} (ours) & \textbf{100.0} & 0.0 & 0.0 & $6.11\pm3.65$ & $0.981\pm0.091$ & $6.02\pm3.55$ \\
\midrule
\multirow{6}{*}{15} & \texttt{ORCA} & 64.0 & 20.0 & 16.0 & $5.22\pm2.81$ & $0.850\pm0.151$ & $3.94\pm2.24$ \\
& \texttt{AVOCADO} & 72.0 & 0.0 & 28.0 & $6.29\pm4.41$ & $0.727\pm0.191$ & $4.64\pm2.42$ \\
& \texttt{VO-polytope} & 52.0 & 47.0 & 1.0 & $7.16\pm3.90$ & $0.531\pm0.065$ & $3.94\pm2.24$ \\
& \texttt{SARL} & 65.0 & 32.0 & 3.0 & $8.38\pm5.64$ & $0.663\pm0.166$ & $5.44\pm3.15$ \\
& \texttt{RL-RVO} & 43.0 & 41.0 & 16.0 & $7.34\pm4.86$ & $0.734\pm0.144$ & $5.43\pm3.31$ \\
\rowcolor[gray]{0.95}
& \texttt{SRL-MPC} (ours) & \textbf{98.0} & 1.0 & 1.0 & $6.58\pm4.00$ & $0.970\pm0.093$ & $6.39\pm3.80$ \\
\midrule
\multirow{6}{*}{20} & \texttt{ORCA} & 29.0 & 41.0 & 30.0 & $5.35\pm3.35$ & $0.816\pm0.173$ & $3.51\pm2.20$ \\
& \texttt{AVOCADO} & 37.0 & 0.0 & 63.0 & $6.69\pm3.98$ & $0.684\pm0.209$ & $4.76\pm2.41$ \\
& \texttt{VO-polytope} & 21.0 & 78.0 & 1.0 & $6.52\pm3.99$ & $0.511\pm0.077$ & $3.51\pm2.20$ \\
& \texttt{SARL} & 38.0 & 62.0 & 0.0 & $8.01\pm5.05$ & $0.653\pm0.162$ & $5.11\pm2.88$ \\
& \texttt{RL-RVO} & 12.0 & 71.0 & 17.0 & $7.53\pm5.79$ & $0.685\pm0.155$ & $5.32\pm3.59$ \\
\rowcolor[gray]{0.95}
& \texttt{SRL-MPC} (ours) & \textbf{93.0} & 6.0 & 1.0 & $7.26\pm4.44$ & $0.960\pm0.080$ & $6.93\pm4.12$ \\
\midrule
\multirow{6}{*}{25} & \texttt{ORCA} & 7.0 & 58.0 & 35.0 & $5.38\pm3.49$ & $0.780\pm0.191$ & $2.84\pm2.00$ \\
& \texttt{AVOCADO} & 11.0 & 0.0 & 89.0 & $7.64\pm5.51$ & $0.638\pm0.228$ & $4.91\pm2.52$ \\
& \texttt{VO-polytope} & 8.0 & 91.0 & 1.0 & $5.38\pm3.68$ & $0.496\pm0.086$ & $2.84\pm2.00$ \\
& \texttt{SARL} & 21.0 & 79.0 & 0.0 & $7.38\pm4.70$ & $0.639\pm0.168$ & $4.75\pm2.81$ \\
& \texttt{RL-RVO} & 1.0 & 90.0 & 9.0 & $6.71\pm5.95$ & $0.657\pm0.162$ & $4.74\pm4.03$ \\
\rowcolor[gray]{0.95}
& \texttt{SRL-MPC} (ours) & \textbf{92.0} & 6.0 & 2.0 & $7.94\pm5.00$ & $0.943\pm0.093$ & $7.41\pm4.41$ \\
\bottomrule
\end{tabular}
}
\resizebox{0.96\linewidth}{!}{\usebox{\baselineTableBox}}
\end{table}

\begin{table}[t!]
\caption{Baseline comparison in out-of-distribution settings. Entries report success$\uparrow$/collision$\downarrow$/timeout$\downarrow$ (\%); $L$ denotes the number of outer iterations.}
\label{tab:ood_results}
\centering
\tiny
\setlength{\tabcolsep}{2.0pt}
\renewcommand{\arraystretch}{0.84}
\resizebox{0.96\linewidth}{!}{
\begin{tabular*}{\wd\baselineTableBox}{@{\extracolsep{\fill}}c|ccccc}
\specialrule{1pt}{1pt}{0pt}
Method & \shortstack{Nonconvex\\union} & \shortstack{Cross\\geometry} &
\shortstack{Circular\\exchange} & \shortstack{Through\\traffic} &
\shortstack{Heterogeneous\\dynamics} \\
\midrule
\texttt{ORCA}~\citep{van2011reciprocal} & 83/10/7 & 71/5/24 & 91/7/2 & 92/1/7 & 88/3/9 \\
\texttt{AVOCADO}~\citep{martinez2025avocado} & 90/2/8 & 61/0/39 & 97/0/3 & \textbf{95}/1/4 & 85/0/15 \\
\texttt{VO-polytope}~\citep{huang2023velocity} & 7/93/0 & 6/94/0 & 1/99/0 & 44/55/1 & 27/73/0 \\
\texttt{SARL}~\citep{chen2019crowd} & 72/4/24 & 44/4/52 & \textbf{100}/0/0 & 44/4/52 & 49/0/51 \\
\texttt{RL-RVO}~\citep{han2022reinforcement} & 45/46/9 & 17/78/5 & 9/91/0 & 37/60/3 & 66/30/4 \\
\rowcolor[gray]{0.95}
\texttt{SRL-MPC} (ours, $L=1$) & \textbf{99}/1/0 & 76/22/2 & 56/44/0 & 85/15/0 & \textbf{97}/1/2 \\
\rowcolor[gray]{0.95}
\texttt{SRL-MPC} (ours, $L=2$) & 96/2/2 & \textbf{87}/13/0 & 83/17/0 & 77/23/0 & 96/2/2 \\
\bottomrule
\end{tabular*}
}
\end{table}
\textbf{Random-polygon setting.}
Table~\ref{tab:baseline_results} shows that \texttt{SRL-MPC} achieves the highest success rate across all robot counts. The gap becomes larger in dense scenes: with $25$ robots, \texttt{SRL-MPC} reaches $92.0\%$ success, while \texttt{ORCA}, \texttt{AVOCADO}, \texttt{VO-polytope}, \texttt{SARL}, and \texttt{RL-RVO} achieve $7.0\%$, $11.0\%$, $8.0\%$, $21.0\%$, and $1.0\%$, respectively. The improvement over the strongest external baseline is $55.0$ percentage points at $20$ robots and $71.0$ percentage points at $25$ robots. \texttt{ORCA} often has the shortest navigation time and path length among arrived robots, but these averages exclude many failed robots in dense scenes. \texttt{AVOCADO} avoids collisions but produces many timeout episodes in dense settings. Although \texttt{VO-polytope} models polygonal geometry more explicitly than circular VO methods, it remains a reactive pairwise velocity-obstacle method. In dense multi-robot scenes, the polygonal velocity cones become tight and overlapping, and the method lacks receding-horizon optimization or adaptive parameter tuning to resolve multi-way conflicts; therefore, its failure mode becomes collision-heavy as density increases. Together, Table~\ref{tab:baseline_results} and Figure~\ref{fig:baseline_comparison} show that \texttt{SRL-MPC} improves the task-completion and safety tradeoff, rather than only changing the speed profile of successful trajectories.

\textbf{Out-of-distribution setting.} Without retraining, we compare the same methods in five OOD settings over $100$ episodes per method and setting (Table~\ref{tab:ood_results}). With one iteration, \texttt{SRL-MPC} success is $99/76/56/85/97\%$ for nonconvex union, cross geometry, circular exchange, through traffic, and heterogeneous dynamics, respectively. Circular-exchange success increases to $83\%$ with two iterations, versus $100\%$ for \texttt{SARL}. The $22/44/15\%$ collision rates in cross, circular, and traffic settings indicate sensitivity to structured interactions.

\textbf{Ablation Study:} To validate the functionality of the proposed components, we also run an ablation study with \texttt{dist\_mpc}, \texttt{manual\_mpc}, \texttt{rule\_mpc}, and \texttt{SRL-MPC}. \texttt{dist\_mpc} uses neither RL nor HOCBF; it replaces the HOCBF term with a static distance-margin penalty over the horizon, $J_{\mathrm{dist}}=(\rho_{\mathrm{obs}}/2)\sum_{j\in\mathcal{N}_i}\sum_{k=1}^{T}[\operatorname{neg}(\bar{H}_{ij,k\mid k})]^2$, where $\bar{H}_{ij,k\mid k}$ is the fixed-geometry distance barrier obtained from the GSFs at step $k$. \texttt{manual\_mpc} uses HOCBF without RL by solving the same HOCBF-MPC problem with fixed handcrafted MPC parameters. \texttt{rule\_mpc} also uses HOCBF without RL, but replaces learned adaptation with a hand coded stuck rule. It uses the default parameters $(w_p,w_u,d_{\mathrm{safe}})=(0.01,10.0,0.30)$ and switches to $(1.0,0.1,0.20)$ when the robot satisfies $|v_i|<0.10\,\mathrm{m/s}$ for $10$ consecutive control steps. \texttt{SRL-MPC} uses both HOCBF and RL based continuous parameter adaptation.
\begin{table}[t!]
\caption{Ablation study in the random convex polygon scenario.}
\centering
\tiny
\setlength{\tabcolsep}{2.0pt}
\renewcommand{\arraystretch}{0.84}
\resizebox{0.98\linewidth}{!}{
\begin{tabular}{lc|ccc|ccc}
\specialrule{1pt}{1pt}{0pt}
\multicolumn{1}{c}{Robots} & \multicolumn{1}{c|}{Method} &
Success$\uparrow$ & Collision$\downarrow$ & Timeout$\downarrow$ &
Time (s)$\downarrow$ & Speed (m/s)$\uparrow$ & Path (m)$\downarrow$ \\
\midrule
10 & \texttt{dist\_mpc} (w/o HOCBF+RL) & 77.0 & 21.0 & 2.0 & $4.99\pm2.49$ & $0.997\pm0.107$ & $5.00\pm2.51$ \\
10 & \texttt{manual\_mpc} (w/o RL) & 96.0 & 1.0 & 3.0 & $6.69\pm5.32$ & $0.983\pm0.091$ & $6.60\pm5.24$ \\
10 & \texttt{rule\_mpc} (rule, w/o RL) & 94.0 & 0.0 & 6.0 & $6.87\pm5.55$ & $0.980\pm0.094$ & $6.76\pm5.43$ \\
\rowcolor[gray]{0.95}
10 & \texttt{SRL-MPC} (ours, with HOCBF+RL) & \textbf{100.0} & 0.0 & 0.0 & $6.11\pm3.65$ & $0.981\pm0.091$ & $6.02\pm3.55$ \\
\midrule
15 & \texttt{dist\_mpc} (w/o HOCBF+RL) & 51.0 & 46.0 & 3.0 & $4.81\pm2.73$ & $0.997\pm0.106$ & $4.81\pm2.74$ \\
15 & \texttt{manual\_mpc} (w/o RL) & 93.0 & 2.0 & 5.0 & $6.85\pm4.74$ & $0.969\pm0.101$ & $6.64\pm4.53$ \\
15 & \texttt{rule\_mpc} (rule, w/o RL) & 92.0 & 1.0 & 7.0 & $7.21\pm5.27$ & $0.965\pm0.098$ & $6.92\pm4.90$ \\
\rowcolor[gray]{0.95}
15 & \texttt{SRL-MPC} (ours, with HOCBF+RL) & \textbf{98.0} & 1.0 & 1.0 & $6.58\pm4.00$ & $0.970\pm0.093$ & $6.39\pm3.80$ \\
\midrule
20 & \texttt{dist\_mpc} (w/o HOCBF+RL) & 15.0 & 84.0 & 1.0 & $3.91\pm2.67$ & $1.003\pm0.097$ & $3.90\pm2.60$ \\
20 & \texttt{manual\_mpc} (w/o RL) & 75.0 & 5.0 & 20.0 & $7.43\pm5.12$ & $0.959\pm0.102$ & $7.13\pm4.85$ \\
20 & \texttt{rule\_mpc} (rule, w/o RL) & 85.0 & 1.0 & 14.0 & $7.93\pm5.30$ & $0.950\pm0.090$ & $7.43\pm4.77$ \\
\rowcolor[gray]{0.95}
20 & \texttt{SRL-MPC} (ours, with HOCBF+RL) & \textbf{93.0} & 6.0 & 1.0 & $7.26\pm4.44$ & $0.960\pm0.080$ & $6.93\pm4.12$ \\
\midrule
25 & \texttt{dist\_mpc} (w/o HOCBF+RL) & 7.0 & 92.0 & 1.0 & $3.57\pm2.70$ & $1.002\pm0.106$ & $3.55\pm2.68$ \\
25 & \texttt{manual\_mpc} (w/o RL) & 68.0 & 13.0 & 19.0 & $7.72\pm5.05$ & $0.945\pm0.110$ & $7.27\pm4.62$ \\
25 & \texttt{rule\_mpc} (rule, w/o RL) & 75.0 & 2.0 & 23.0 & $8.67\pm5.97$ & $0.931\pm0.103$ & $7.94\pm5.22$ \\
\rowcolor[gray]{0.95}
25 & \texttt{SRL-MPC} (ours, with HOCBF+RL) & \textbf{92.0} & 6.0 & 2.0 & $7.94\pm5.00$ & $0.943\pm0.093$ & $7.41\pm4.41$ \\
\bottomrule
\end{tabular}
}
\label{tab:ablation_results}
\end{table}
Table~\ref{tab:ablation_results} shows the superior performance of \texttt{SRL-MPC} over the ablation baselines. \texttt{manual\_mpc} is competitive at low and moderate densities, but drops to $75.0\%$ and $68.0\%$ success at $20$ and $25$ robots, whereas \texttt{SRL-MPC} reaches $93.0\%$ and $92.0\%$. \texttt{rule\_mpc} improves over \texttt{manual\_mpc} at high density by using a binary stuck rule: at $20$ and $25$ robots, success rates increase from $75.0\%$ to $85.0\%$ and from $68.0\%$ to $75.0\%$, respectively. However, \texttt{rule\_mpc} still depends on hand tuned thresholds and only switches between two parameter settings. Its wider default safety distance also makes it prone to timeouts, with $23.0\%$ timeout at $25$ robots. In contrast, \texttt{SRL-MPC} improves over \texttt{rule\_mpc} by $6.0$, $6.0$, $8.0$, and $17.0$ percentage points for $10$, $15$, $20$, and $25$ robots, respectively, by adapting parameters continuously from local GSFs. At $25$ robots, \texttt{rule\_mpc} remains collision conservative but timeout heavy, whereas \texttt{SRL-MPC} reaches $92.0\%$ success by reducing timeout to $2.0\%$ while keeping collision to $6.0\%$. \texttt{dist\_mpc} is fast among the robots that arrive, but its collision rate increases to $84.0\%$ at $20$ robots and $92.0\%$ at $25$ robots, confirming that the static distance margin penalty is insufficient as the primary safety mechanism in dense polygonal crowds. Unlike the static distance margin penalty, the HOCBF residual couples barrier values across consecutive predicted steps, so it penalizes not only instantaneous distance violations but also unsafe trends along the horizon. This provides more informative optimization guidance in dense interactions. The comparison isolates the benefits of the HOCBF residual, rule based adaptation, and learned continuous parameter adaptation in sequence.

\textbf{Robustness Analysis:}
We further evaluate \texttt{SRL-MPC} under perception noise, action delay, and parameter delay in the $N=15$ random polygon setting, using the same scenario, evaluation seed, and max-step setting as the main evaluation. The success rate remains $100.0\%$ with Gaussian neighbor-position noise of $\sigma=0.02\,\mathrm{m}$ and $83.0\%$ at $\sigma=0.05\,\mathrm{m}$, and reaches $99.0\%$ with one-step action delay ($0.1\,\mathrm{s}$). The method is also insensitive to delayed RL parameter updates: with a five-step parameter delay ($0.5\,\mathrm{s}$), success remains $100.0\%$. Larger perception noise or action delay causes a clear collision increase, indicating that accurate short-horizon state estimation and low-latency actuation remain important. The detailed sweep is reported in Table~\ref{tab:robustness_results}. We also deploy \texttt{SRL-MPC} on real-world robot platforms to validate practical effectiveness and real-time execution, as shown in Figure~\ref{fig:real_world_experiment} and the supplementary video.

\begin{table}[t]
\caption{Robustness analysis under perception noise and delay perturbations for $N=15$.}
\centering
\scriptsize
\setlength{\tabcolsep}{0pt}
\renewcommand{\arraystretch}{0.70}
\begin{tabular*}{0.98\linewidth}{@{\extracolsep{\fill}}cccccccccccc@{}}
\toprule
\multicolumn{4}{c}{Perception noise} &
\multicolumn{4}{c}{Action delay} &
\multicolumn{4}{c}{Parameter delay} \\
\cmidrule(lr){1-4}\cmidrule(lr){5-8}\cmidrule(lr){9-12}
$\sigma$ (m) & Succ. & Coll. & T.out &
$K$ (step) & Succ. & Coll. & T.out &
$K$ (step) & Succ. & Coll. & T.out \\
\midrule
0.00 & 98.0 & 1.0 & 1.0 & 0 & 98.0 & 1.0 & 1.0 & 0 & 98.0 & 1.0 & 1.0 \\
0.02 & 100.0 & 0.0 & 0.0 & 1 & 99.0 & 1.0 & 0.0 & 1 & 98.0 & 2.0 & 0.0 \\
0.05 & 83.0 & 17.0 & 0.0 & 2 & 82.0 & 17.0 & 1.0 & 2 & 100.0 & 0.0 & 0.0 \\
0.10 & 40.0 & 60.0 & 0.0 & 3 & 2.0 & 98.0 & 0.0 & 3 & 99.0 & 1.0 & 0.0 \\
0.20 & 1.0 & 99.0 & 0.0 & 5 & 0.0 & 100.0 & 0.0 & 5 & 100.0 & 0.0 & 0.0 \\
\bottomrule
\end{tabular*}
\label{tab:robustness_results}
\end{table}

\textbf{Additional evaluations.}
\textit{(i) Outer-iteration ablation} (Appendix~\ref{app:additional_iterations}). In the additional $20$-robot sweep, one to four iterations increase \texttt{SRL-MPC} success from $94\%$ to $100\%$ and reduce collisions from $4\%$ to $0\%$, while mean controller time increases from $9.500$ to $36.703$ ms per robot and step. At four iterations, \texttt{manual\_mpc} also has zero observed collisions but $71\%$ timeouts. The appendix reports the complete sweeps and separate residual diagnostics; performance is not monotonic in iteration count.
\textit{(ii) Training-seed variation} (Appendix~\ref{app:training_seed_variation}). Three additional PPO training seeds yield $91.0\pm2.6\%$ and $89.7\pm0.6\%$ success at $20$ and $25$ robots, respectively. Table~\ref{tab:training_seeds} reports the dense and OOD results, with sample standard deviations across trained policies.
\textit{(iii) Hardware evaluation} (Appendix~\ref{app:hardware_evaluation}). Six robots are evaluated over $20$ paired trials per method (Table~\ref{tab:hardware_results}). \texttt{SRL-MPC} achieves $85\%$ success and $15\%$ collisions, versus $60\%$ and $40\%$ for \texttt{ORCA}. The robots share the same physical chassis; different footprint sizes are configured in the controllers. Navigation time and speed are computed over successful trials.
\textit{(iv) Computational cost} (Appendix~\ref{app:additional_compute}). We report GSF backend timings, a module breakdown, and comparisons with RDA and an online OBCA adapter. 

\section{Conclusion}
\label{sec:conclusion}
This paper presents \texttt{SRL-MPC}, a shape-aware reinforcement learned MPC framework for collision avoidance in crowded dynamic environments with multiple robots and obstacles. It represents geometric constraints through GSFs based on support function transformation, formulates degree-2 HOCBF constraints in the coupled primal problem, and decomposes online planning into a GSF update and a local HOCBF-MPC subproblem where the decomposed HOCBF residual is softly penalized. Reinforcement learning adapts MPC parameters from neighboring shape-aware geometric features, while the executed control remains the solution of an explicit MPC optimization problem.
Experiments in random polygon scenarios with density sweeps show that \texttt{SRL-MPC} achieves the highest success rate among representative baselines, especially in dense scenes where the baselines degrade sharply. The HOCBF residual improves empirical collision avoidance, while learned parameter adaptation primarily reduces timeouts and improves task progress. Future work includes uncertainty-aware neighbor prediction, richer nonconvex body decompositions, and real-world deployment with onboard sensing and computation.

\begin{ack}
This work is supported by the National Natural Science Foundation of China (No. 62422606), Hong Kong Research Grant Council General Research Fund (No. 17213925), Shenzhen Loop Area Institute Fund (No. FPF10120250006), and HKU Shanghai Intelligent Computing Research Center. This work is also partially supported by the National Natural Science Foundation of China (Grant No. 62371444), and the Shenzhen Science and Technology Program (Grant No. RCYX20231211090206005, JCYJ20241202124934046).
\end{ack}

\bibliographystyle{plainnat}
\bibliography{reference/Thesis}
\clearpage
\appendix

\section{Technical appendices and supplementary material}
\label{app:technical_appendices}

\subsection{Kinematic and Geometric Model}
\label{app:model_details}
This appendix provides the detailed definitions of the feasible set $\mathcal{F}_i$ and the occupied set $\mathbb{Z}_i(\mathbf{s}_{i,k})$ used in Section~\ref{sec:problem}. The state and control satisfy
\begin{equation}
\mathbf{s}_{i,k+1}=f_i(\mathbf{s}_{i,k},\mathbf{u}_{i,k}),\quad
k=0,\ldots,T-1 ,
    \label{eq:kinematics}
\end{equation}
where $f_i:\mathcal{X}_i\times\mathcal{V}_i\rightarrow\mathbb{R}^{n}$ is the discrete-time robot dynamics, $\mathcal{X}_i$ is the state space, and $\mathcal{V}_i$ is the admissible input set. In planar navigation, $n=3$ and $\mathbf{s}_{i,k}=[x_{i,k},y_{i,k},\theta_{i,k}]^\top$. The implementation uses a differential-drive model with $\mathbf{u}_{i,k}=[v_{i,k},\omega_{i,k}]^\top$ and sampling time $\Delta t$:
\begin{equation}
\begin{aligned}
    x_{i,k+1}&=x_{i,k}+\Delta t\,v_{i,k}\cos\theta_{i,k},\quad
    y_{i,k+1}=y_{i,k}+\Delta t\,v_{i,k}\sin\theta_{i,k},\quad
    \theta_{i,k+1}=\theta_{i,k}+\Delta t\,\omega_{i,k}.
\end{aligned}
\label{eq:app_diff_drive}
\end{equation}
The nonlinear kinematics are linearized around the nominal trajectory as in~\citep{han2023rda}. The input and input increment satisfy $\mathbf{u}_{i,\min}\le\mathbf{u}_{i,k}\le\mathbf{u}_{i,\max}$ and $\Delta\mathbf{u}_{i,\min}\le\Delta\mathbf{u}_{i,k}\le\Delta\mathbf{u}_{i,\max}$ for $k=0,\ldots,T-1$, where $\Delta\mathbf{u}_{i,k}=\mathbf{u}_{i,k}-\mathbf{u}_{i,k-1}$ and $\mathbf{u}_{i,-1}$ is the previously applied control. Thus, $\mathcal{F}_i$ is the set of all $(\mathcal{S}_i,\mathcal{U}_i)$ satisfying~\eqref{eq:kinematics}, the input constraints, and the initial condition $\mathbf{s}_{i,0}=\mathbf{s}_i^{\mathrm{cur}}$.

Following the geometric convention in~\citep{zhang2020optimization, han2025neupan}, the convex body-frame occupied set of object $i$ and its world-frame transformation at prediction step $k$ are
\begin{equation}
\begin{aligned}
    \mathbb{C}_i
    &= \{\,\mathbf{z}\in\mathbb{R}^2 \mid \mathbf{G}_i\mathbf{z}\preceq_{\mathcal{K}_i}\mathbf{h}_i\,\},
    \quad
    \mathbb{Z}_i(\mathbf{s}_{i,k})
    = \{\,\mathbf{R}_{i,k}\mathbf{z}+\mathbf{p}_{i,k} \mid \mathbf{z}\in\mathbb{C}_i\,\}.
\end{aligned}
    \label{eq:shape}
\end{equation}
Here, $\mathbf{z}$ is a body-frame point, $\mathbf{G}_i$ and $\mathbf{h}_i$ define the conic inequalities, $\mathcal{K}_i$ is a proper cone, and $\preceq_{\mathcal{K}_i}$ denotes the partial order induced by $\mathcal{K}_i$. The matrix $\mathbf{R}_{i,k}\triangleq\mathbf{R}(\theta_{i,k})\in SO(2)$ is the planar rotation matrix, where
\begin{equation}
    \mathbf{R}(\theta)
    =
    \begin{bmatrix}
        \cos\theta & -\sin\theta\\
        \sin\theta & \cos\theta
    \end{bmatrix}.
\label{eq:app_rotation_matrix}
\end{equation}
For a nominal neighbor state $\bar{\mathbf{s}}_{j,k}$, the corresponding position and rotation are denoted by $\bar{\mathbf{p}}_{j,k}$ and $\bar{\mathbf{R}}_{j,k}$.
The support function used in~\eqref{eq:dual_distance} is defined as
\begin{equation}
    \sigma_{\mathbb{C}}(\boldsymbol{\xi})
    =
    \sup\{\,\boldsymbol{\xi}^{\top}\mathbf{z}\mid\mathbf{z}\in\mathbb{C}\,\},
    \quad
    \sigma_{\mathbb{C}_{\mathrm{cir}}}(\boldsymbol{\xi})
    =
    r\lVert\boldsymbol{\xi}\rVert_2,
    \quad
    \sigma_{\mathbb{C}_{\mathrm{poly}}}(\boldsymbol{\xi})
    =
    \max_{a=1,\ldots,n_v}(\mathbf{v}^{a})^\top\boldsymbol{\xi},
    \label{eq:app_support_function_examples}
\end{equation}
where $\mathbb{C}_{\mathrm{cir}}$ is a body-centered circle with radius $r$, and $\mathbb{C}_{\mathrm{poly}}$ is a polygon with body-frame vertices $\{\mathbf{v}^{a}\}_{a=1}^{n_v}$.
Substituting~\eqref{eq:shape} into~\eqref{eq:collision_avoidance} gives the equivalent body-frame distance program
\begin{equation}
\begin{aligned}
    D_{ij,k}
    &=
    \min_{\mathbf{z}_i,\mathbf{z}_j}
    \Big\lVert
    \mathbf{R}_{i,k}\mathbf{z}_i+\mathbf{p}_{i,k}
    -
    \big(\bar{\mathbf{R}}_{j,k}\mathbf{z}_j+\bar{\mathbf{p}}_{j,k}\big)
    \Big\rVert_2 \\
    \mathrm{s.t.}\quad
    & \mathbf{G}_i\mathbf{z}_i\preceq_{\mathcal{K}_i}\mathbf{h}_i,
    \quad
    \mathbf{G}_j\mathbf{z}_j\preceq_{\mathcal{K}_j}\mathbf{h}_j .
\end{aligned}
\label{eq:app_body_frame_distance}
\end{equation}

\subsection{Limitations}
\label{app:limitations}
The main limitation is the computational cost of solving the local HOCBF-MPC subproblem. As shown in Table~\ref{tab:compute_cost}, \texttt{SRL-MPC} is heavier than lightweight reactive baselines because each control step solves a convex MPC problem after the geometric feature update. Nevertheless, the measured mean per-robot controller time remains at the $10\,\mathrm{ms}$ level in the tested dense scenarios: $5.79$, $6.73$, $9.37$, and $10.34\,\mathrm{ms}$ for $10$, $15$, $20$, and $25$ robots, respectively, supporting real-time per-robot control in the platform experiment shown in Figure~\ref{fig:real_world_experiment}. 

\subsection{Broad Impact}
\label{app:broad_impact}
This work aims to improve the safety and reliability of autonomous robot navigation in shared spaces, including warehouses, service-robot environments, and heterogeneous robot fleets. By keeping the executed control inside an explicit MPC optimization problem and using RL to adapt interpretable parameters, the framework may reduce the risk of opaque end-to-end policy behavior in dense navigation tasks. Potential negative impacts mainly come from premature deployment: inaccurate state estimation, model mismatch, computation delays, or unmodeled pedestrian behavior could still lead to unsafe motion.

\subsection{Additional Challenging Scenarios}
\label{app:additional_scenarios}
\begin{figure}[t]
    \centering
    \scriptsize
    \setlength{\tabcolsep}{1.5pt}
    \renewcommand{\arraystretch}{0.82}
    \begin{tabular}{@{}p{0.12\linewidth}ccccc@{}}
        Cross geometry &
        \includegraphics[width=0.155\textwidth]{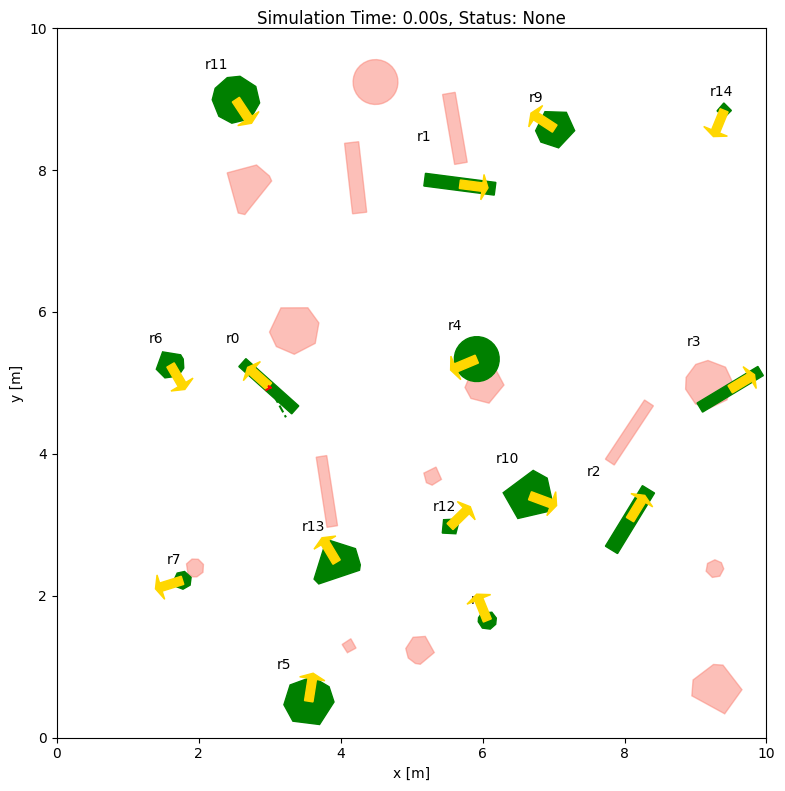} &
        \includegraphics[width=0.155\textwidth]{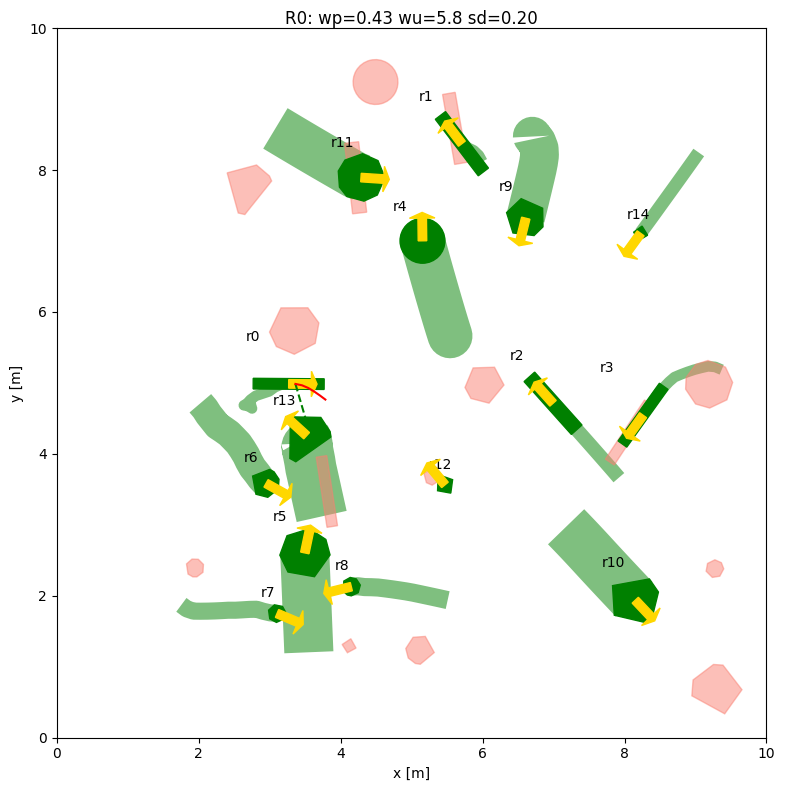} &
        \includegraphics[width=0.155\textwidth]{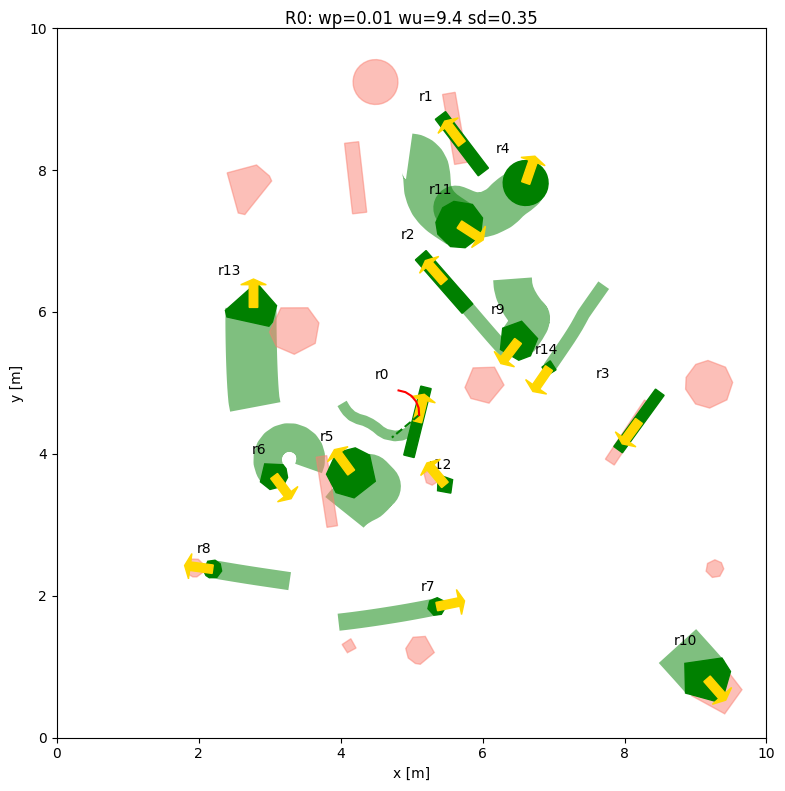} &
        \includegraphics[width=0.155\textwidth]{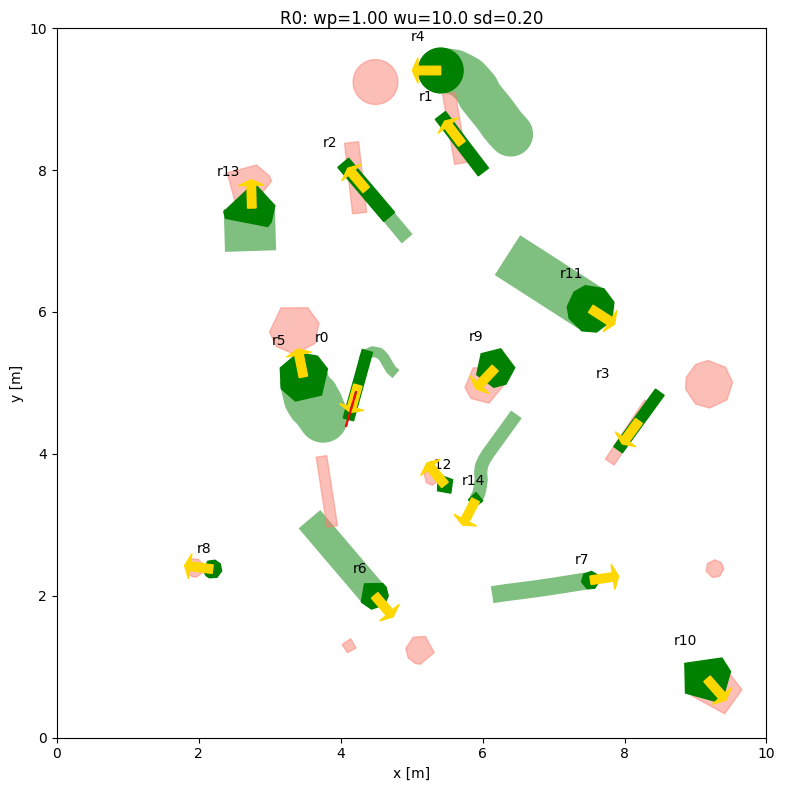} &
        \includegraphics[width=0.155\textwidth]{extra_cross_step090} \\
        & Step 0 & Step 22 & Step 45 & Step 67 & Step 90 \\
        \shortstack[l]{Nonconvex\\union} &
        \includegraphics[width=0.155\textwidth]{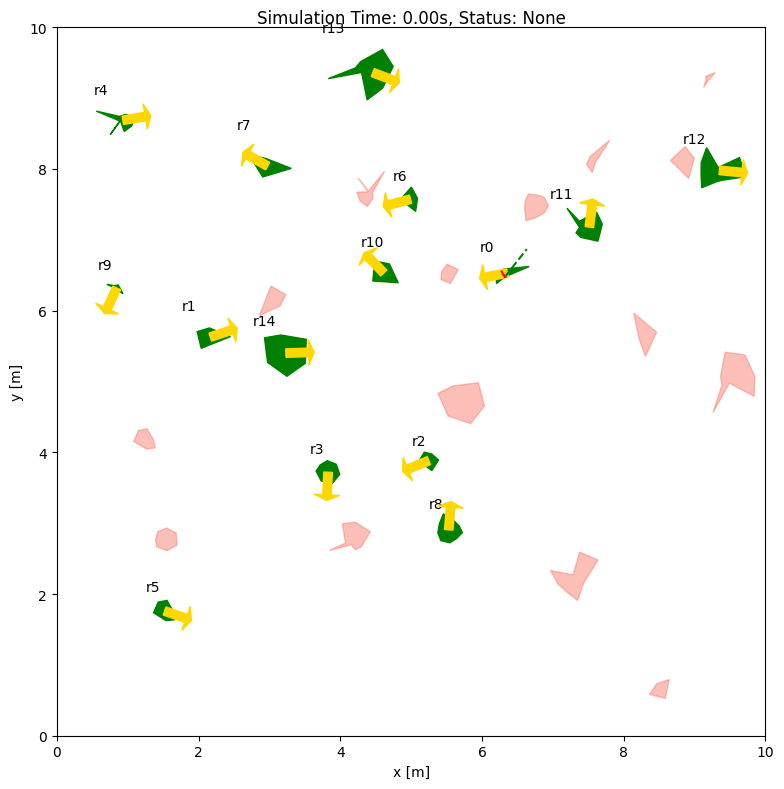} &
        \includegraphics[width=0.155\textwidth]{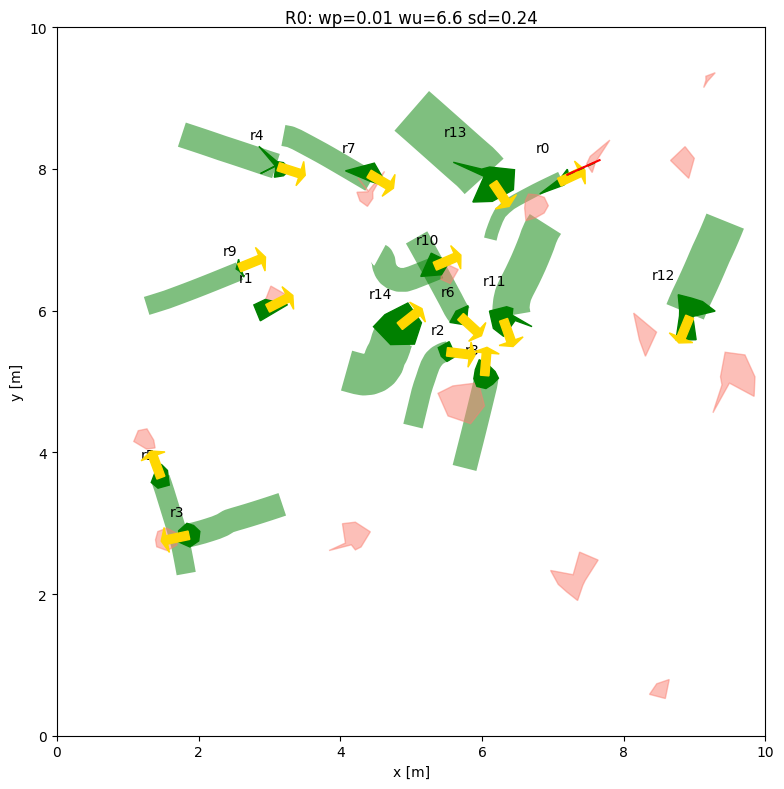} &
        \includegraphics[width=0.155\textwidth]{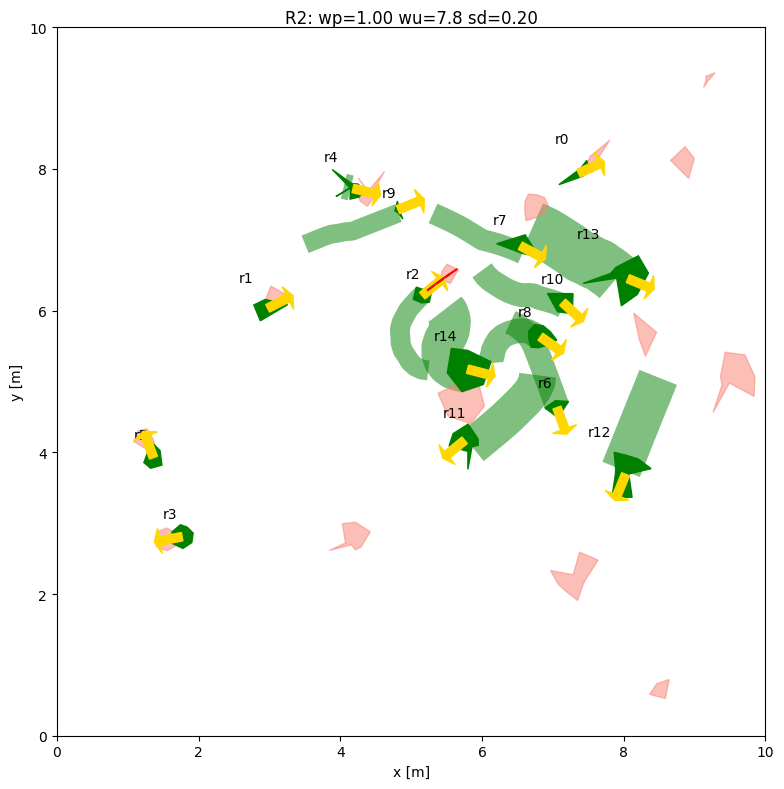} &
        \includegraphics[width=0.155\textwidth]{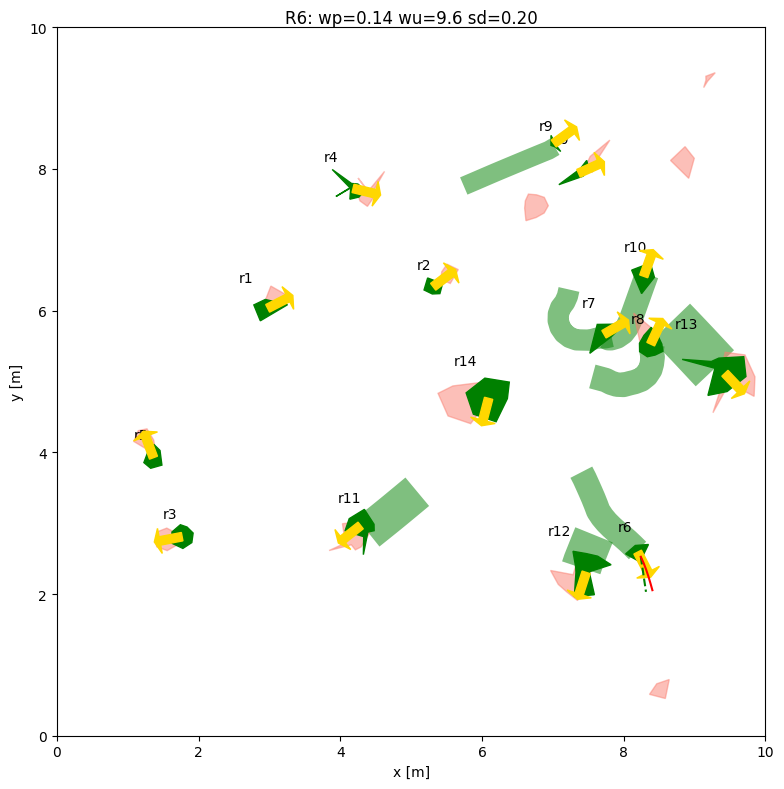} &
        \includegraphics[width=0.155\textwidth]{extra_nonconvex_step096} \\
        & Step 0 & Step 24 & Step 48 & Step 72 & Step 96 \\
        Circular polygon &
        \includegraphics[width=0.155\textwidth]{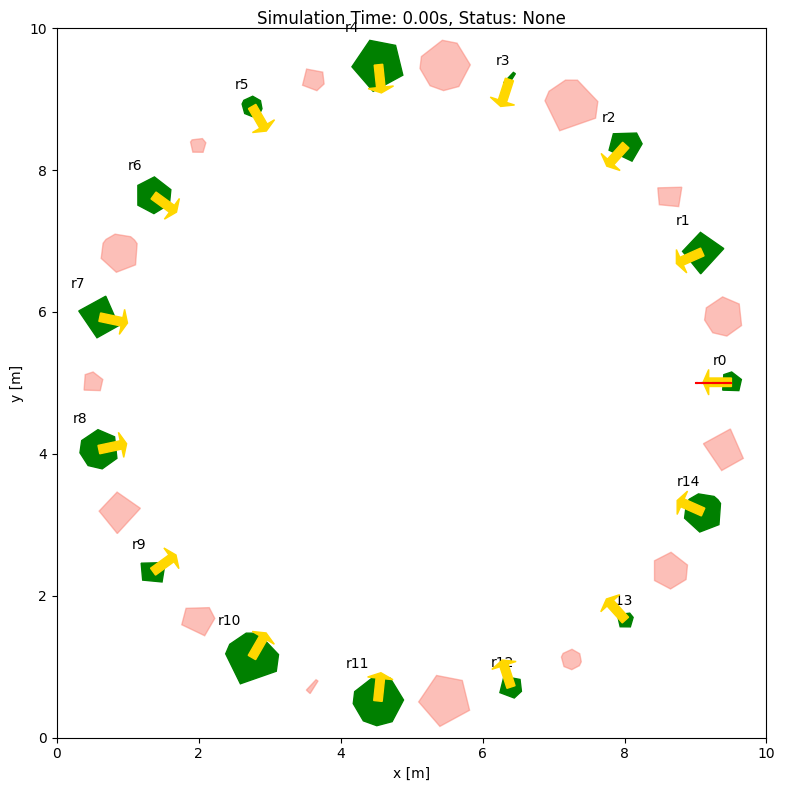} &
        \includegraphics[width=0.155\textwidth]{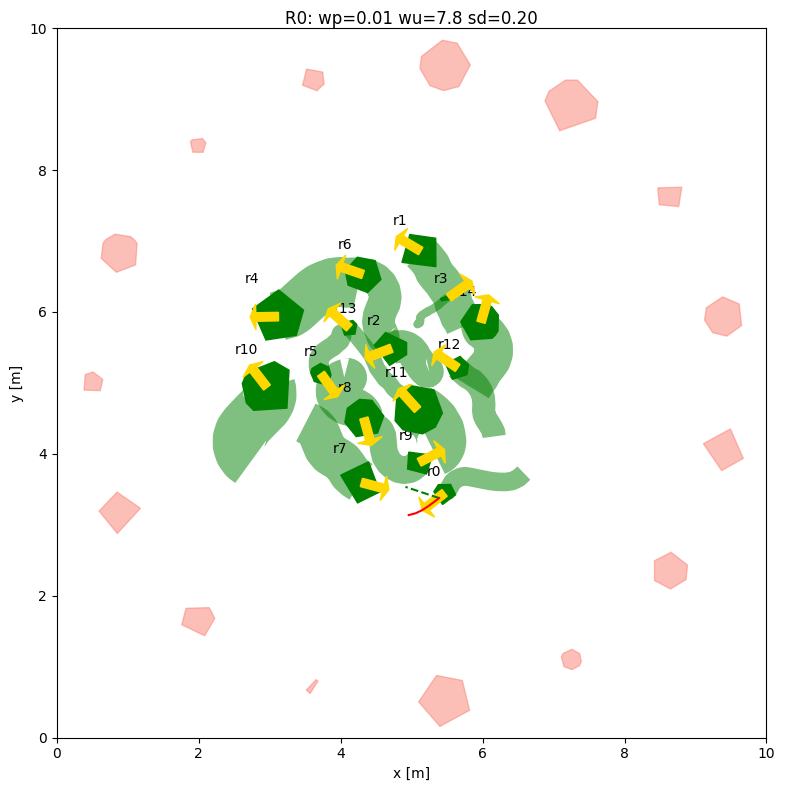} &
        \includegraphics[width=0.155\textwidth]{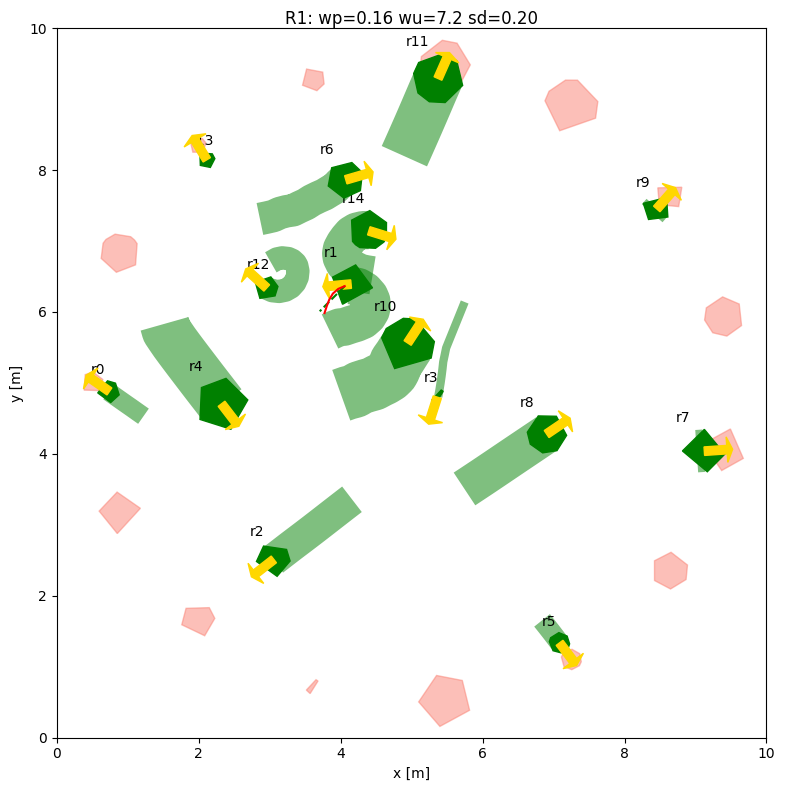} &
        \includegraphics[width=0.155\textwidth]{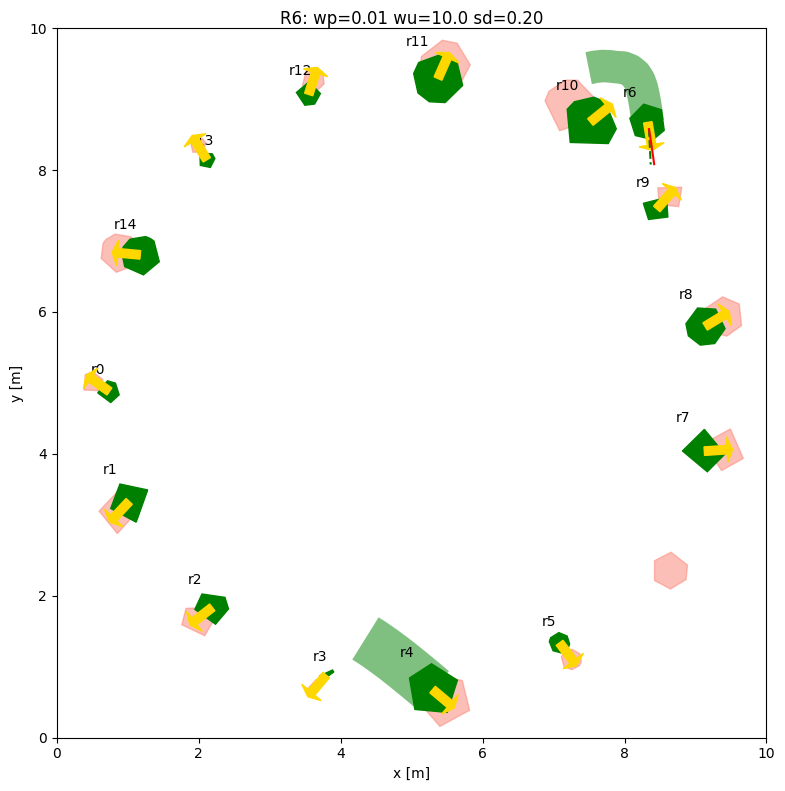} &
        \includegraphics[width=0.155\textwidth]{extra_circular_step250} \\
        & Step 0 & Step 62 & Step 125 & Step 188 & Step 250 \\
        Through traffic &
        \includegraphics[width=0.155\textwidth]{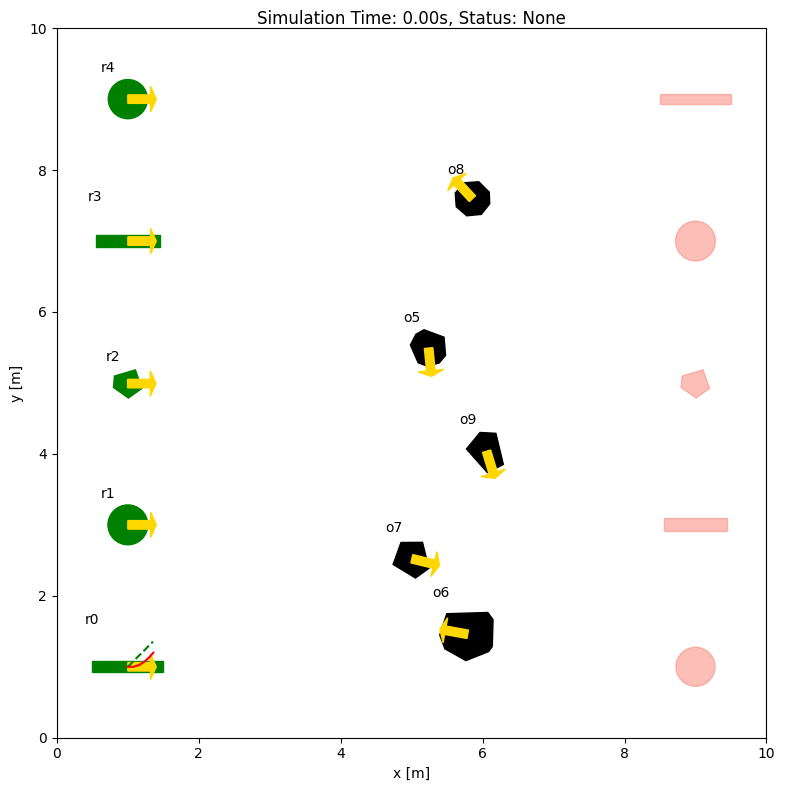} &
        \includegraphics[width=0.155\textwidth]{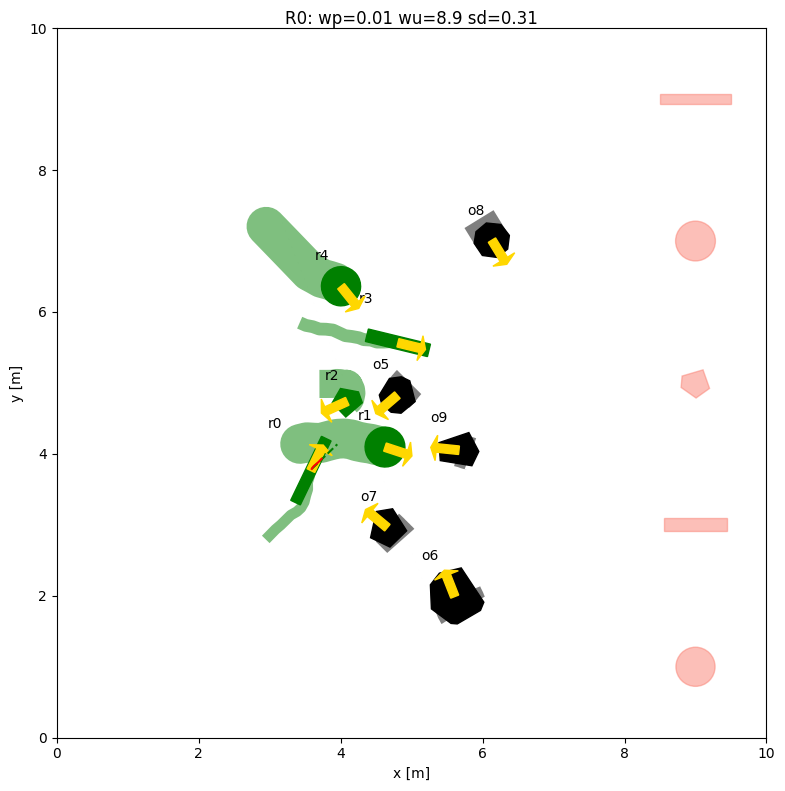} &
        \includegraphics[width=0.155\textwidth]{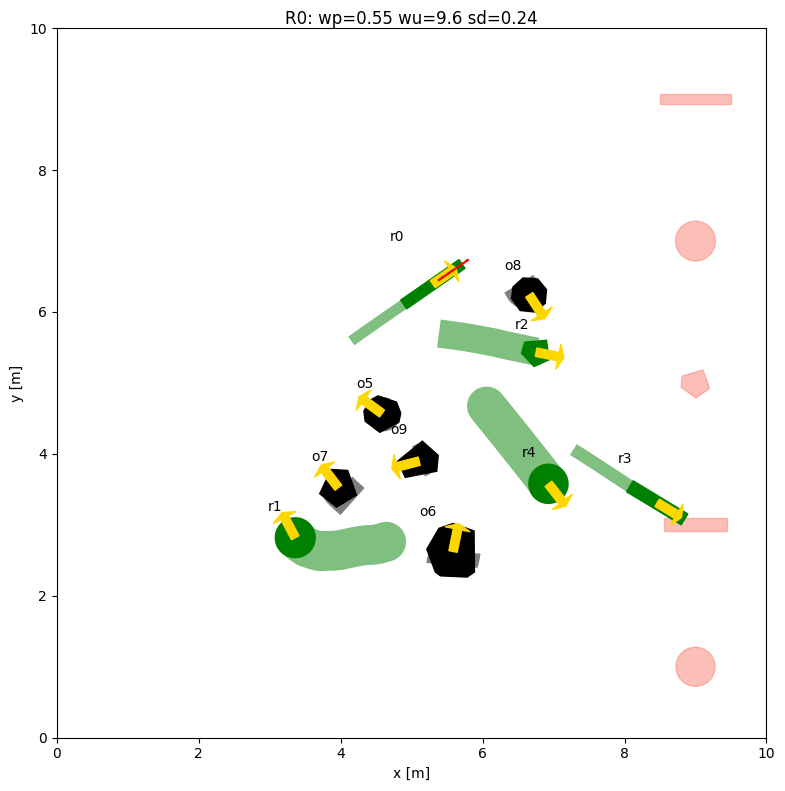} &
        \includegraphics[width=0.155\textwidth]{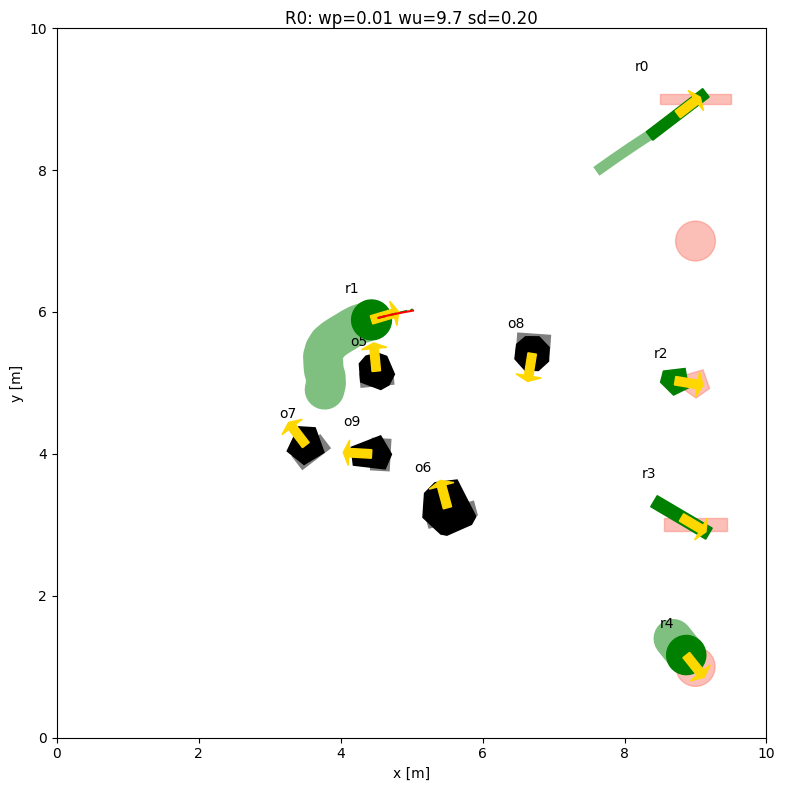} &
        \includegraphics[width=0.155\textwidth]{extra_traffic_step170} \\
        & Step 0 & Step 42 & Step 85 & Step 127 & Step 170
    \end{tabular}
    \caption{Additional challenging scenario rollouts. Rows show cross geometry, nonconvex union, circular polygon, and through traffic scenarios sampled from held-out seeds.}
    \label{fig:app_extra_combined_scenarios}
\end{figure}

Figure~\ref{fig:app_extra_combined_scenarios} shows additional held-out scenarios, including cross geometry, nonconvex union, circular polygon, and through traffic layouts. The snapshots are sampled from one episode per scenario from the beginning to the final step, illustrating that \texttt{SRL-MPC} can guide polygonal robots through different geometry layouts while adapting the MPC parameters during the rollout. Section~\ref{app:reactive_parameter_adaptation} further discusses a reactive-pair example.

\subsubsection{Reactive Parameter Adaptation}
\label{app:reactive_parameter_adaptation}
\begin{figure}[t]
    \centering
    \includegraphics[width=0.92\linewidth]{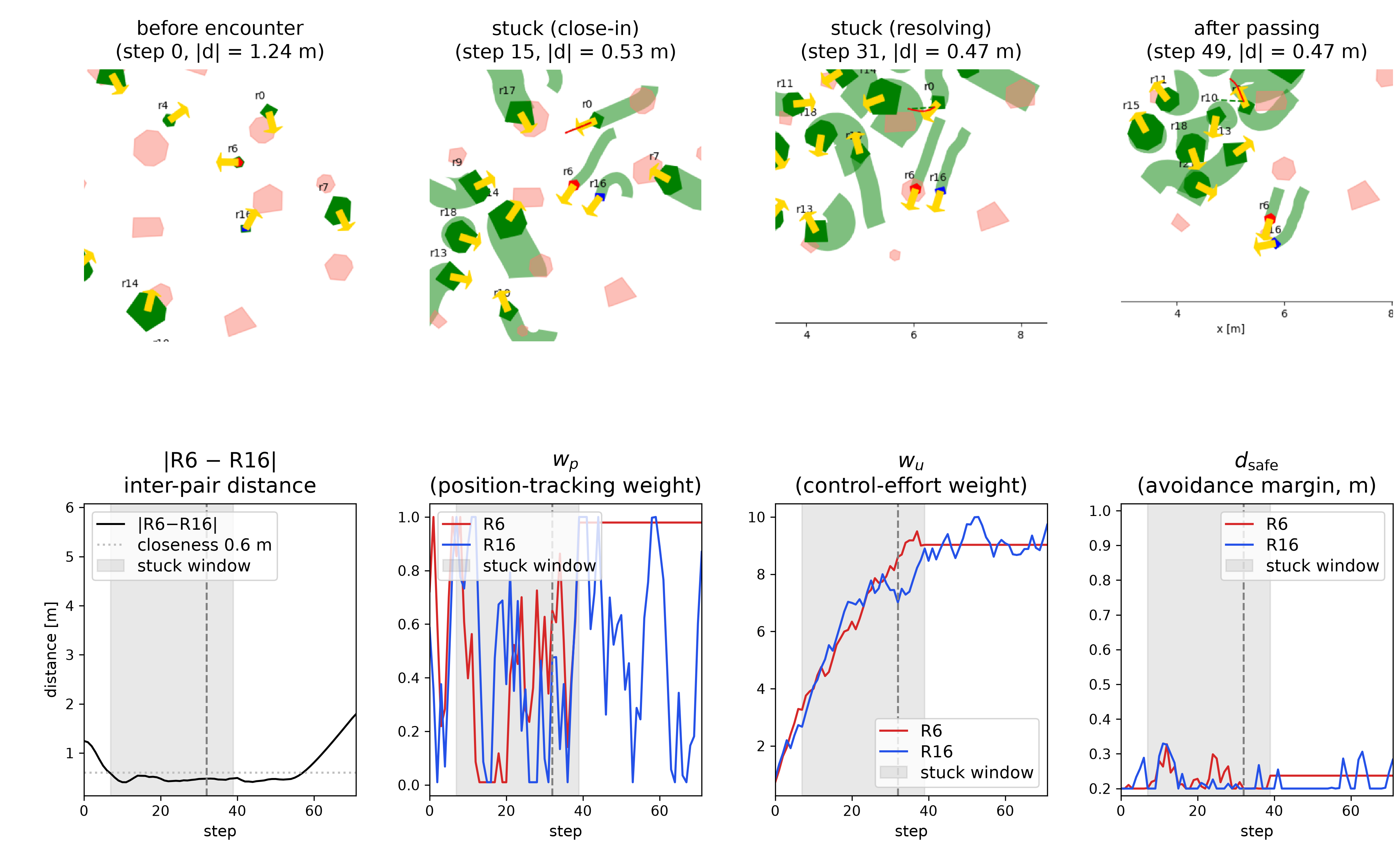}
    \caption{Reactive pair example in the random polygon scenario. The upper row shows the interaction between R6 and R16 at representative time steps, and the lower row shows the corresponding changes in inter-robot distance, $w_p$, $w_u$, and $d_{\mathrm{safe}}$, indicating that the learned policy adapts the MPC parameters to the current neighbor geometry.}
    \label{fig:app_reactive_pair_parameters}
\end{figure}

Figure~\ref{fig:app_reactive_pair_parameters} visualizes a close interaction between two robots and the corresponding learned parameter changes. As the neighboring geometry changes during the interaction, the policy reacts by updating $w_p$, $w_u$, and $d_{\mathrm{safe}}$ online. This example validates that the learned adaptation is not a fixed parameter schedule, but responds to the current GSFs induced by nearby robot shapes and distances.

\subsection{Implementation and Experiment Parameters}
\label{app:experiment_parameters}
This subsection summarizes the main implementation settings used for the reported experiments. Table~\ref{tab:controller_parameters} lists the controller and scenario parameters, Table~\ref{tab:rl_parameters} lists the reinforcement learning and neural-network parameters, and Table~\ref{tab:evaluation_parameters} lists the evaluation and baseline settings. The notation follows the main text: $T$ is the MPC horizon, $K$ is the number of neighbor GSFs in $\mathbf{o}_{i,\mathrm{neighbor}}^t$, $H_c$ is the HOCBF horizon, and $\lambda_i^t=[w_{p,i}^t,w_{u,i}^t,d_{\mathrm{safe},i}^t]^\top$ is the adaptive parameter vector.

\begin{table}[t]
\caption{Controller and scenario parameters used by \texttt{SRL-MPC}.}
\centering
\scriptsize
\setlength{\tabcolsep}{3pt}
\renewcommand{\arraystretch}{1.05}
\begin{tabular}{p{0.17\linewidth}p{0.22\linewidth}p{0.52\linewidth}}
\toprule
Group & Symbol or item & Value \\
\midrule
Prediction model
& $T$, $\Delta t$
& $5$, $0.1\,\mathrm{s}$ \\
HOCBF-MPC
& $K$, $|\mathcal{N}_i^{\mathrm{cbf}}|$, $H_c$, distance horizon, $\gamma$, $\rho_{\mathrm{obs}}$
& $5$, $3$, $3$, $1$, $0.9$, $100$ \\
Adaptive parameters
& $\lambda_i^t$
& Initial: $(1.0,1.0,0.2\,\mathrm{m})$; $\Lambda=[0.01,1.0]\times[0.1,10.0]\times[0.2,1.0]$; $w_\theta=0.01$ fixed \\
Solver and geometry
& $Q_1$, $Q_2$
& \texttt{Shapely}, Splitting Conic Solver (\texttt{SCS}), one outer iteration; distance cost disabled for \texttt{SRL-MPC} \\
Training world
& $N_{\mathrm{train}}$, workspace, $\mathbb{C}_i$
& $15$ robots; $10\,\mathrm{m}\times10\,\mathrm{m}$; convex polygon radius $[0.1,0.4]\,\mathrm{m}$; irregularity $[0.1,1.0]$; goal threshold $0.3\,\mathrm{m}$ \\
\bottomrule
\end{tabular}
\label{tab:controller_parameters}
\end{table}

\begin{table}[t]
\caption{Reinforcement learning and neural-network parameters.}
\centering
\scriptsize
\setlength{\tabcolsep}{3pt}
\renewcommand{\arraystretch}{1.05}
\begin{tabular}{p{0.17\linewidth}p{0.22\linewidth}p{0.52\linewidth}}
\toprule
Group & Symbol or item & Value \\
\midrule
Observation
& $\mathbf{o}_i^t$, $\mathbf{o}_{i,\mathrm{neighbor}}^t$
& $79=4+3\times5\times5$ \\
Action
& $\Delta\lambda_i^t$
& $[-0.5,0.5]$, $[-0.5,0.5]$, $[-0.1,0.1]\,\mathrm{m}$; absolute values clipped to $\Lambda$ \\
Network
& CNN, actor, critic
& Convolution $3\!\rightarrow\!16\!\rightarrow\!32$; actor $36\rightarrow64\rightarrow6$; critic $36\rightarrow64\rightarrow1$; log-scale clamp $[-5,0]$ \\
PPO schedule
& Frames, batch, mini-batch, epochs
& $10^7$, $1000$, $500$, $3$ \\
PPO and generalized advantage estimation (GAE) parameters
& $\gamma_{\mathrm{PPO}}$, $\lambda_{\mathrm{GAE}}$
& $0.99$, $0.95$; clip $0.2$; entropy $0.01$; learning rate $3\times10^{-4}$; max grad norm $1.0$ \\
Training reward
& $r_{\mathrm{arr}}$, $r_{\mathrm{col}}$, $r_{\mathrm{safe}}$, $r_{\mathrm{step}}$
& $40$, $80$, $1$, $0.05$ \\
Safety reward
& $d_{\mathrm{safe}}^{\min}$, $d_{\mathrm{safe}}^{\max}$, $\epsilon_d$
& $0.2\,\mathrm{m}$, $1.0\,\mathrm{m}$, $0.01\,\mathrm{m}$ \\
\bottomrule
\end{tabular}
\label{tab:rl_parameters}
\end{table}

\begin{table}[t]
\caption{Evaluation and baseline settings.}
\centering
\scriptsize
\setlength{\tabcolsep}{3pt}
\renewcommand{\arraystretch}{1.05}
\begin{tabular}{p{0.17\linewidth}p{0.22\linewidth}p{0.52\linewidth}}
\toprule
Group & Symbol or item & Value \\
\midrule
Evaluation sweep
& $N$, $n_{\mathrm{ep}}$, $E_{\max}$
& $N\in\{10,15,20,25\}$; $n_{\mathrm{ep}}=100$; $E_{\max}=500$ steps ($50\,\mathrm{s}$) \\
Randomization
& Train/test seeds
& Training seed $0$; evaluation seed $100$; disjoint scenario sequences \\
Scenario
& \texttt{random\_polygon}
& Random start-goal pairs; random convex polygon footprints; collision mode \texttt{stop} \\
Ablation methods
& \texttt{dist\_mpc}, \texttt{manual\_mpc}, \texttt{rule\_mpc}
& Fixed $w_p=0.01$, $w_\theta=0.01$, $w_u=10.0$; \texttt{rule\_mpc} uses $d_{\mathrm{safe}}=0.30$ by default and switches to $(w_p,w_u,d_{\mathrm{safe}})=(1.0,0.1,0.20)$ after $10$ stuck steps \\
VO baselines
& \texttt{ORCA}, \texttt{AVOCADO}, \texttt{VO-polytope}
& Neighbor distance $15\,\mathrm{m}$; time horizon $0.5\,\mathrm{s}$; safety buffer $0.1\,\mathrm{m}$; \texttt{VO-polytope} neighbor distance $3\,\mathrm{m}$ \\
RL baselines
& \texttt{RL-RVO}, \texttt{SARL}
& Pretrained policies; desired speed $1.0\,\mathrm{m/s}$; $\Delta t=0.1\,\mathrm{s}$ \\
\bottomrule
\end{tabular}
\label{tab:evaluation_parameters}
\end{table}

\subsection{Compute-Cost Profile}
\label{app:compute_cost}
Using the controller, learning, and evaluation settings summarized in Tables~\ref{tab:controller_parameters},~\ref{tab:rl_parameters}, and~\ref{tab:evaluation_parameters}, we profile the per-robot per-step controller time in the \texttt{random\_polygon} scenario using three episodes per cell, a maximum of $50$ steps per episode, and seed $100$. Each sample is the wall-clock duration of one method decision call divided by the number of robots. Table~\ref{tab:compute_cost} reports mean and standard deviation over timed steps for the robot counts used in the main evaluation.

\begin{table}[t]
\caption{Per-robot per-step controller compute time in milliseconds.}
\centering
\scriptsize
\setlength{\tabcolsep}{3pt}
\renewcommand{\arraystretch}{1.05}
\resizebox{0.98\linewidth}{!}{
\begin{tabular}{lcccc}
\toprule
Method & $N=10$ & $N=15$ & $N=20$ & $N=25$ \\
\midrule
\texttt{manual\_mpc} & $5.482\pm1.585$ & $6.586\pm1.715$ & $8.488\pm1.418$ & $9.955\pm1.803$ \\
\texttt{dist\_mpc} & $5.202\pm1.449$ & $7.313\pm1.125$ & $8.693\pm1.688$ & $11.260\pm1.358$ \\
\texttt{ORCA} & $0.021\pm0.003$ & $0.022\pm0.001$ & $0.024\pm0.002$ & $0.026\pm0.002$ \\
\texttt{AVOCADO} & $0.022\pm0.002$ & $0.026\pm0.002$ & $0.028\pm0.002$ & $0.028\pm0.001$ \\
\texttt{SARL} & $15.612\pm0.904$ & $16.353\pm1.810$ & $17.346\pm1.670$ & $18.561\pm1.737$ \\
\texttt{RL-RVO} & $0.357\pm0.048$ & $0.366\pm0.037$ & $0.370\pm0.033$ & $0.377\pm0.041$ \\
\rowcolor[gray]{0.95}
\texttt{SRL-MPC} (ours) & $5.786\pm2.023$ & $6.729\pm1.669$ & $9.365\pm1.679$ & $10.340\pm1.801$ \\
\bottomrule
\end{tabular}
}
\label{tab:compute_cost}
\end{table}

\begingroup
\subsection{Out-of-Distribution Evaluation Details}
\label{app:ood_quantitative}
Table~\ref{tab:ood_results} evaluates the same trained \texttt{SRL-MPC} policy without retraining in five settings: nonconvex union footprints; cross geometry with elongated rectangles, circles, and convex polygons; circular exchange; through traffic with five controlled robots and five nonreciprocal moving obstacles; and heterogeneous dynamics with five differential-drive polygon robots and five omnidirectional circular robots. Each method and setting uses $100$ episodes, evaluation base seed $100$, and at most $500$ control steps. The OOD protocol derives paired scene seeds independently for each episode. The one- and two-iteration \texttt{SRL-MPC} runs use the same checkpoint and paired seeds.
For nonconvex geometry, the implementation computes separation directions from the original geometry while using convex-hull-equivalent support values. Structured interactions remain challenging: \texttt{SARL} performs best in circular exchange, while \texttt{ORCA} and \texttt{AVOCADO} have higher success and fewer collisions in through traffic. Increasing the iteration count improves cross and circular exchange outcomes but reduces success in the other three settings.

\subsection{Outer-Iteration Ablation and Residual Diagnostics}
\label{app:additional_iterations}
Table~\ref{tab:outer_iterations} compares outer-iteration counts in the random-polygon setting. Each outcome cell contains $100$ episodes with base seed $100$ and at most $500$ steps. Within each method and robot count, only the iteration count changes and episode seeds are paired. Controller times come from separate profiles with $150$ sampled control steps per cell, except the $25$-robot, two-iteration cell, which has $141$ usable records; each step's controller time is normalized by the robot count. Increasing iterations can reduce observed collisions while increasing computation and timeouts. With $20$ robots and four iterations, \texttt{manual\_mpc} and \texttt{SRL-MPC} both have zero observed collisions, but their timeout rates are $71\%$ and $0\%$, respectively. \texttt{SRL-MPC} success decreases again at five iterations, so navigation performance is not monotonic in iteration count.

\begin{table}[t]
\centering
\small
\caption{Outer-iteration ablation. Outcome rates are percentages; time is mean controller latency in ms per robot and control step.}
\label{tab:outer_iterations}
\begin{tabular*}{\linewidth}{@{\extracolsep{\fill}}rlrrrrr@{}}
\toprule
Robots & Method & $L$ & Success & Collision & Timeout & Time \\
\midrule
20 & \texttt{manual\_mpc} & 1 & 81 & 6 & 13 & 8.723 \\
20 & \texttt{manual\_mpc} & 2 & 46 & 2 & 52 & 17.551 \\
20 & \texttt{manual\_mpc} & 3 & 41 & 1 & 58 & 25.944 \\
20 & \texttt{manual\_mpc} & 4 & 29 & 0 & 71 & 34.011 \\
20 & \texttt{manual\_mpc} & 5 & 41 & 0 & 59 & 41.670 \\
\midrule
20 & \texttt{SRL-MPC} & 1 & 94 & 4 & 2 & 9.500 \\
20 & \texttt{SRL-MPC} & 2 & 99 & 1 & 0 & 18.464 \\
20 & \texttt{SRL-MPC} & 3 & 96 & 0 & 4 & 26.884 \\
20 & \texttt{SRL-MPC} & 4 & 100 & 0 & 0 & 36.703 \\
20 & \texttt{SRL-MPC} & 5 & 94 & 4 & 2 & 45.706 \\
\midrule
25 & \texttt{SRL-MPC} & 1 & 89 & 8 & 3 & 11.728 \\
25 & \texttt{SRL-MPC} & 2 & 89 & 4 & 7 & 24.661 \\
25 & \texttt{SRL-MPC} & 3 & 94 & 3 & 3 & 34.360 \\
25 & \texttt{SRL-MPC} & 4 & 91 & 2 & 7 & 41.797 \\
25 & \texttt{SRL-MPC} & 5 & 94 & 0 & 6 & 57.573 \\
\bottomrule
\end{tabular*}
\end{table}

\subsection{Variation Across PPO Training Seeds}
\label{app:training_seed_variation}
Table~\ref{tab:training_seeds} reports the additional evaluation across three independently trained policies, using PPO training seeds $1$, $2$, and $3$. Values are means and sample standard deviations across policies, rather than across individual episodes.

\begin{table}[t]
\centering
\small
\caption{Performance across three PPO training seeds. Entries are mean $\pm$ sample standard deviation in percentage points.}
\label{tab:training_seeds}
\begin{tabular*}{\linewidth}{@{\extracolsep{\fill}}lccc@{}}
\toprule
Setting & Success (\%) & Collision (\%) & Timeout (\%) \\
\midrule
Random polygon, $N=20$ & $91.0\pm2.6$ & $7.3\pm2.5$ & $1.7\pm1.2$ \\
Random polygon, $N=25$ & $89.7\pm0.6$ & $8.3\pm2.1$ & $2.0\pm1.7$ \\
Nonconvex union & $97.7\pm0.6$ & $2.3\pm0.6$ & $0.0\pm0.0$ \\
Cross geometry & $77.3\pm1.8$ & $20.7\pm1.6$ & $2.0\pm1.0$ \\
Circular exchange & $56.0\pm0.0$ & $44.0\pm0.0$ & $0.0\pm0.0$ \\
Through traffic & $90.7\pm1.5$ & $9.0\pm1.7$ & $0.3\pm0.6$ \\
Heterogeneous dynamics & $95.0\pm1.7$ & $0.3\pm0.6$ & $4.7\pm2.1$ \\
\bottomrule
\end{tabular*}
\end{table}

\subsection{Quantitative Hardware Evaluation}
\label{app:hardware_evaluation}
Six robots navigate in a shared workspace over $20$ paired trials per method, for $40$ trials in total. Corresponding trials use the same start positions and goals sampled with the same seed. Navigation time and average speed are computed over successful trials. Table~\ref{tab:hardware_results} reports success, collision, navigation, and controller-computation metrics. The $25$-percentage-point observed success improvement is descriptive; a two-sided exact McNemar test yields $p=0.125$. The physical robots have the same chassis and footprint; heterogeneous footprint sizes are configured in their controllers. This tests deployment with different configured geometry parameters, while validation on physically heterogeneous robots remains future work.

\begin{table}[t]
\centering
\small
\caption{Hardware results with $20$ trials per method. Controller time is in ms per robot and step.}
\label{tab:hardware_results}
\begin{tabular*}{\linewidth}{@{\extracolsep{\fill}}lrrrrr@{}}
\toprule
Method & \shortstack{Success\\(\%)} & \shortstack{Collision\\(\%)} &
\shortstack{Navigation\\time (s)} & \shortstack{Speed\\(m/s)} &
\shortstack{Controller\\time (ms)} \\
\midrule
\texttt{ORCA} & 60 & 40 & $21\pm3.14$ & $0.43\pm0.09$ & 0.078 \\
\texttt{SRL-MPC} & 85 & 15 & $17\pm2.15$ & $0.48\pm0.12$ & 8.53 \\
\bottomrule
\end{tabular*}
\end{table}

\subsection{Computation Analysis}
\label{app:additional_compute}
We compare the cost of computing the same GSF using GEOS ShortestLine and a dual support-function program implemented with CvxpyLayer/SCS. Table~\ref{tab:gsf_backend_latency} uses batches of $120$ queries, corresponding to $24$ neighbors over five horizon steps before neighbor selection in the $25$-robot setting. Timings exclude geometry generation, optimization-layer construction, and warm-up. Both backends pass the benchmark's numerical agreement checks on separated convex polygons. This isolates GSF computation rather than measuring an entire MPC controller. Table~\ref{tab:module_timing} gives the mean module breakdown for one \texttt{SRL-MPC} iteration with $20$ robots. The separately measured controller total can differ slightly from the sum of rounded module values.

\begin{table}[t]
\centering
\small
\caption{GSF backend latency for $120$ queries. Times are medians in ms per batch. The last column is the median of paired optimization/GEOS timing ratios.}
\label{tab:gsf_backend_latency}
\begin{tabular*}{\linewidth}{@{\extracolsep{\fill}}rrrr@{}}
\toprule
Edges & GEOS & CvxpyLayer/SCS & Median paired ratio \\
\midrule
3 & 0.102 & 11.139 & $108.37\times$ \\
4 & 0.111 & 13.066 & $119.34\times$ \\
8 & 0.142 & 35.225 & $263.94\times$ \\
16 & 0.196 & 127.758 & $641.92\times$ \\
32 & 0.343 & 550.835 & $1560.72\times$ \\
\bottomrule
\end{tabular*}
\end{table}

\begin{table}[t]
\centering
\small
\caption{One-iteration \texttt{SRL-MPC} module profile with $20$ robots: mean ms per robot and control step over $150$ sampled steps.}
\label{tab:module_timing}
\begin{tabular*}{\linewidth}{@{\extracolsep{\fill}}lr@{}}
\toprule
Module & Time \\
\midrule
RL policy and parameter application & 0.058 \\
Path update & 0.206 \\
Nominal trajectory update & 0.019 \\
$Q_1$: GSF computation & 5.994 \\
$Q_2$: MPC optimization & 3.188 \\
Other operations & 0.034 \\
\midrule
Total & 9.500 \\
\bottomrule
\end{tabular*}
\end{table}
\endgroup

\end{document}